\documentclass[sigconf,nonacm]{acmart}

\usepackage{amsmath,amsfonts}
\usepackage{algorithmic}
\usepackage{graphicx}
\usepackage{textcomp}
\usepackage{xcolor}
\usepackage{subfigure}
\usepackage{algorithm}
\usepackage{algorithmic}
\usepackage{romannum}

\usepackage{amsthm}
\newtheorem{theorem}{Theorem}

\newtheorem{remark}[theorem]{Remark}

\AtBeginDocument{%
  \providecommand\BibTeX{{%
    \normalfont B\kern-0.5em{\scshape i\kern-0.25em b}\kern-0.8em\TeX}}}

\begin{document}

\title{Effective Multi-User Delay-Constrained Scheduling with\\Deep Recurrent Reinforcement Learning
}


\author{Pihe Hu, Ling Pan, Yu Chen, Zhixuan Fang, Longbo Huang}
\affiliation{%
  \institution{Tsinghua University}
  \state{Beijing}
  \country{China}}
\email{{hph19, pl17, c-y19}@mails.tsinghua.edu.cn, {zfang, longbohuang}@tsinghua.edu.cn}

\renewcommand{\shortauthors}{Pihe Hu, Ling Pan, Yu Chen, Zhixuan Fang and Longbo Huang.}

\begin{abstract}
    Multi-user delay constrained scheduling is important in many real-world applications including wireless communication, live streaming, and cloud computing. Yet, it poses a critical challenge since the scheduler needs to make real-time decisions to guarantee the delay and resource constraints simultaneously without prior information of system dynamics, which can be time-varying and hard to estimate.
    Moreover, many practical scenarios suffer from partial observability issues, e.g., due to sensing noise or hidden correlation. 
    To tackle these challenges, we propose a deep reinforcement learning (DRL) algorithm, named \underline{R}ecurrent \underline{S}oftmax \underline{D}elayed \underline{D}eep \underline{D}ouble \underline{D}eterministic Policy Gradient ($\mathtt{RSD4}$)\footnote{Code available at https://github.com/hupihe/RSD4}, which is a data-driven method based on a Partially Observed Markov Decision Process (POMDP) formulation. $\mathtt{RSD4}$ guarantees resource and delay constraints by Lagrangian dual and delay-sensitive queues, respectively.
    It also efficiently tackles partial observability with a memory mechanism enabled by the recurrent neural network (RNN) and introduces user-level decomposition and node-level merging to ensure scalability.
    Extensive experiments on simulated/real-world datasets demonstrate that $\mathtt{RSD4}$ is robust to system dynamics and partially observable environments, and achieves superior performances over existing DRL and non-DRL-based methods.
    
\end{abstract}




\maketitle
    
    \vspace{-.2cm}
	\section{Introduction}
	
	Due to the emergence of many real-time interactive applications, e.g., online games, virtual reality (VR), and cloud computing, and the increasingly more rigid user requirements,  delay-constrained scheduling has become a central problem in guaranteeing the satisfying quality of experience in many areas. 
	For example, express delivery is a typical delay-sensitive scheduling problem.
	According to \cite{ma2017fast}, a small increase in the delivery time will significantly impact customers’ perceived ambiguity and risk, and reduce satisfaction. 
	Delay-constrained scheduling is also critical for data communications \cite{zhang2019reles}, video streaming \cite{bhattacharyya2019qflow}, and data center management \cite{qu2017reliability}.
	
	However, delay-constrained scheduling is challenging in many aspects.
    First, the scheduler needs to satisfy latency and resource constraints, while the delay metric depends on the overall dynamics and control of the system across time, and the resource constraint further couples scheduling decisions. 
    Second, system dynamics, e.g., user channels in mobile networks, are hard to trace since distributions of underlying random components can be highly dynamic and correlated.
    Third, practical scheduling systems usually face a large number of users and have complex network structures, and demand highly scalable solutions for large-scale and multi-hop scenarios. 
    For example, live video platforms such as YouTube and Instagram have thousands of millions of daily active users  \cite{dao2022contemporary}. 
    Fourth, due to sensing noise and hidden correlation, practical systems can also suffer from partial observability issues, e.g., most IoT devices cannot have perfect knowledge of a dynamic channel environment due to hardware limitation and short sensing time  \cite{xie2020dynamic}, and channel states in network systems can also be hard to obtain  \cite{li2013network}. %
    This universality of partial observability in real-world scheduling problems demands highly robust algorithms.
    
\begin{figure}[htbp]
            \vspace{-.4cm}
            \centering\includegraphics[width=.6\columnwidth]{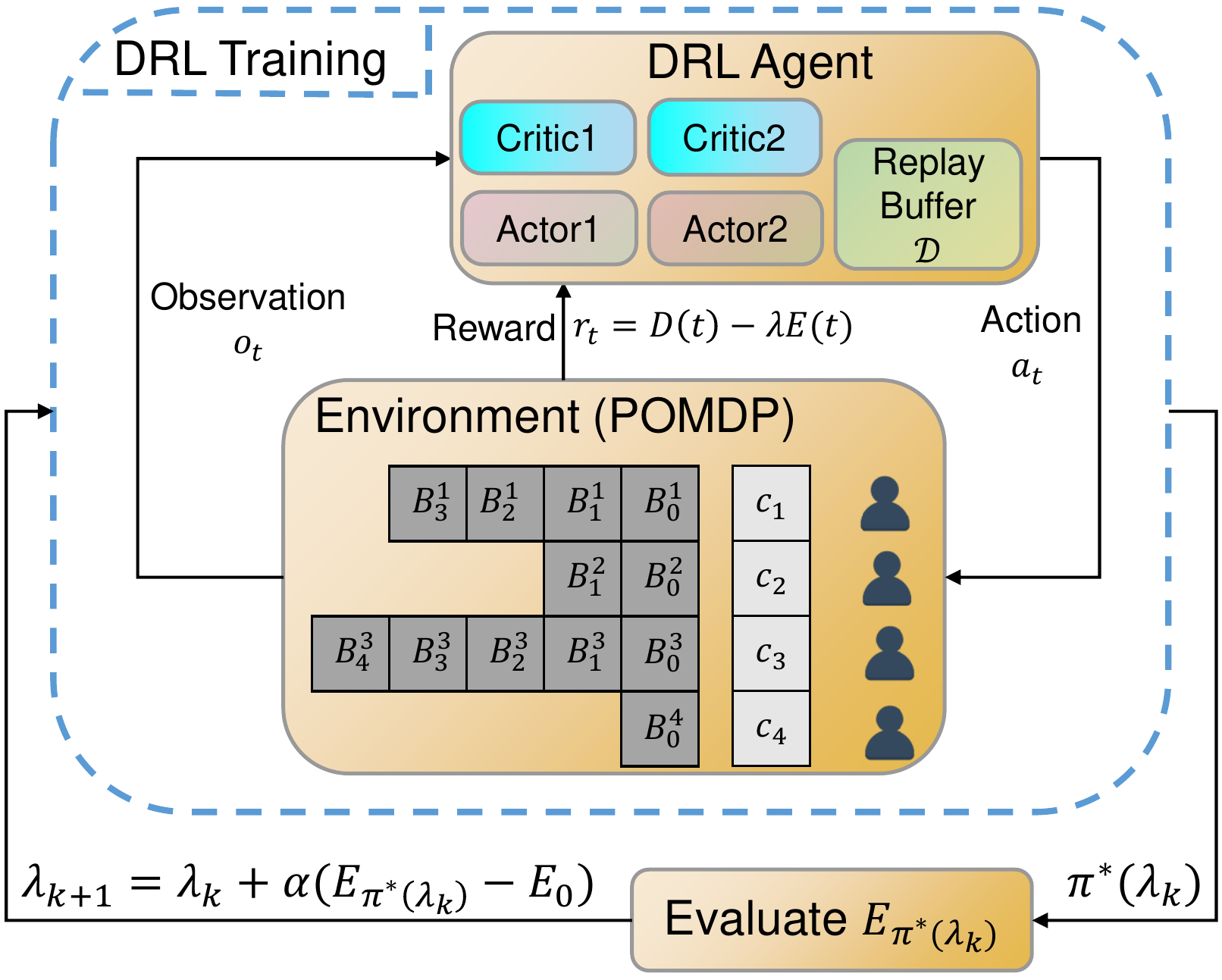}
    	\vspace{-.2cm}
    	\caption{$\mathtt{RSD4}$ framework for delay-constrained scheduling.}
    	\label{fig:framework}
    	\vspace{-.5cm}
    \end{figure}

    Many algorithms have been proposed for scheduling problems based on different methods, including queueing-based methods, e.g.,  \cite{huang2015backpressure, scully2019simple}, optimization-based methods, e.g.,  \cite{hou2010utility, li2019optimal}, dynamic programming (DP)-based algorithms, e.g., \cite{singh2018throughput, chen2018timely}, and Lyapunov control-based approaches, e.g., \cite{pan2018energy, lv2021contract}.  
	However, these approaches either require prior knowledge about system dynamics, or suffer from curse-of-dimensionality due to large state space, or focus on stability constraints rather than delay. 
	To overcome the above limitations, we propose a deep reinforcement learning (DRL)-based algorithm, named \underline{R}ecurrent \underline{S}oftmax \underline{D}elayed \underline{D}eep \underline{D}ouble \underline{D}eterministic Policy Gradient ($\mathtt{RSD4}$), whose procedure is depicted in Figure \ref{fig:framework}. $\mathtt{RSD4}$ builds upon the recurrent deterministic policy gradient \cite{heess2015memory} and softmax deterministic policy gradient \cite{pan2020softmax}, and introduces several novel components for handling the scheduling problem in partially observed settings. 
	Specifically, $\mathtt{RSD4}$ is an end-to-end method based on a partially observed Markov decision process (POMDP) formulation with a Lagrange dual update. It does not require any prior knowledge of system dynamics and effectively captures hidden system correlation across time slots by a recurrent module. 
	$\mathtt{RSD4}$ makes use of a softmax operator to improve value estimation in training and robustness in handling complex system dynamics. Finally, it introduces unified training for improving training efficiency, and user-level decomposition and node-level merging to support large-scale  and multi-hop scenarios.

	The $\mathtt{RSD4}$ framework has several unique features that make it suitable for handling delay-sensitive scheduling tasks, especially in partially observable systems. 
    Firstly, its base POMDP formulation is general to capture the partial observability issue that often arises in real-world systems. 
    Secondly, the proposed $\mathtt{RSD4}$ algorithm possesses key advantages of existing DRL algorithms, i.e., being a data-driven and end-to-end method that does not require prior knowledge of the system dynamics. 
    Thirdly, the $\mathtt{RSD4}$ algorithm is highly scalable, making it suitable for practical implementation in large-scale resource-constrained scenarios, where the system observation can be partial and even not instantaneous. 
    We conduct extensive experiments to validate the performance of $\mathtt{RSD4}$ on both simulated environments and real-world datasets. The results demonstrate that $\mathtt{RSD4}$ significantly outperforms classical non-DRL methods and existing DRL benchmarks in various scenarios.
    Our framework and results shed light on designing scalable and efficient DRL algorithms for scheduling complex systems.

    The main contributions of this paper are summarized as follows.
	
		(\romannumeral1) We formulate a general POMDP  framework for investigating large-scale multi-user delay-constrained scheduling problems with average resource constraint, which is suitable for partially observable systems, and provides two novel functions for scalability, i.e., user-level decomposition and node-level merging.

		(\romannumeral2) We propose a novel DRL-based algorithm, $\mathtt{RSD4}$, to tackle the partial observability problem and achieve robust value prediction.
		$\mathtt{RSD4}$ builds upon the recurrent policy gradient method and softmax-based value estimation. It introduces a unified training approach to improve training efficiency. 
		It also adopts a double-branch neural network architecture and delayed policy update to learn a robust policy against different system dynamics. 
		
		(\romannumeral3) We conduct extensive experiments on real datasets, showing that $\mathtt{RSD4}$ efficiently learns system dynamics under both large-scale and multi-hop cases, and significantly outperforms existing scheduling methods, especially in various partially-observable settings. 
	
	\vspace{-.2cm}
	\section{Related Work}

	Many existing works have studied the  scheduling problem.
	Among the many techniques adopted, four methodologies have received much attention, including queueing theory-based method, e.g.,  \cite{huang2015backpressure, scully2019simple}, optimization-based method, e.g.,  \cite{hou2010utility, li2019optimal}, dynamic programming-based control, e.g., \cite{singh2018throughput, chen2018timely}, Lyapunov-based optimization, e.g., \cite{pan2018energy, lv2021contract}.
	However, these approaches either do not explicitly capture the delay constraint or require precise system dynamics which can be highly non-trivial in real systems.
	Besides, DP-based algorithms often suffer from the curse-of-dimensionality and  are not scalable to large-scale scenarios. 
	DRL has been receiving much attention in the scheduling field, due to its  generalization  and scalability capability, and has been adopted in several scheduling scenarios, e.g., video streaming \cite{mao2017neural}, Multipath TCP control \cite{zhang2019reles}, network reconfigurability \cite{bhattacharyya2019qflow}, MAC scheduling \cite{moon2020neuro}, and resource-constrained scheduling, e.g., \cite{nasir2019multi, meng2020power, xiao2017reinforcement, he2019joint}. 
	However, the aforementioned  works either do not ensure average resource constraints or require a perfectly observable system state as input.
	In addition, they do not  consider large-scale or multi-hop systems.
	
	Our proposed $\mathtt{RSD4}$ is a data-driven end-to-end algorithm, which well handles the partial observability issue under the POMDP formulation.
	Besides, the delay and resource constraint are handled by delay-sensitive queues and the Lagrangian dual, respectively. 
	It also adopts user-level decomposition and node-level merging to significantly extend the scalability.
    
    \vspace{-.2cm}
    \section{Problem and Preliminary}
    In this section, we present our scheduling problem description. 
    For ease of presentation, we focus on the  single-hop scheduling problem in Section \ref{subsec:sm} and the corresponding Lagrangian dual in Section \ref{subsec:ld}. 
    The multi-hop problem will be presented later in Section \ref{subsec:multihop}.
    
    \vspace{-.2cm}
	\subsection{The Scheduling Problem}\label{subsec:sm}
\begin{figure}[htbp]
		\centering
		\vspace{-.2cm}
		\includegraphics[width=.6\columnwidth]{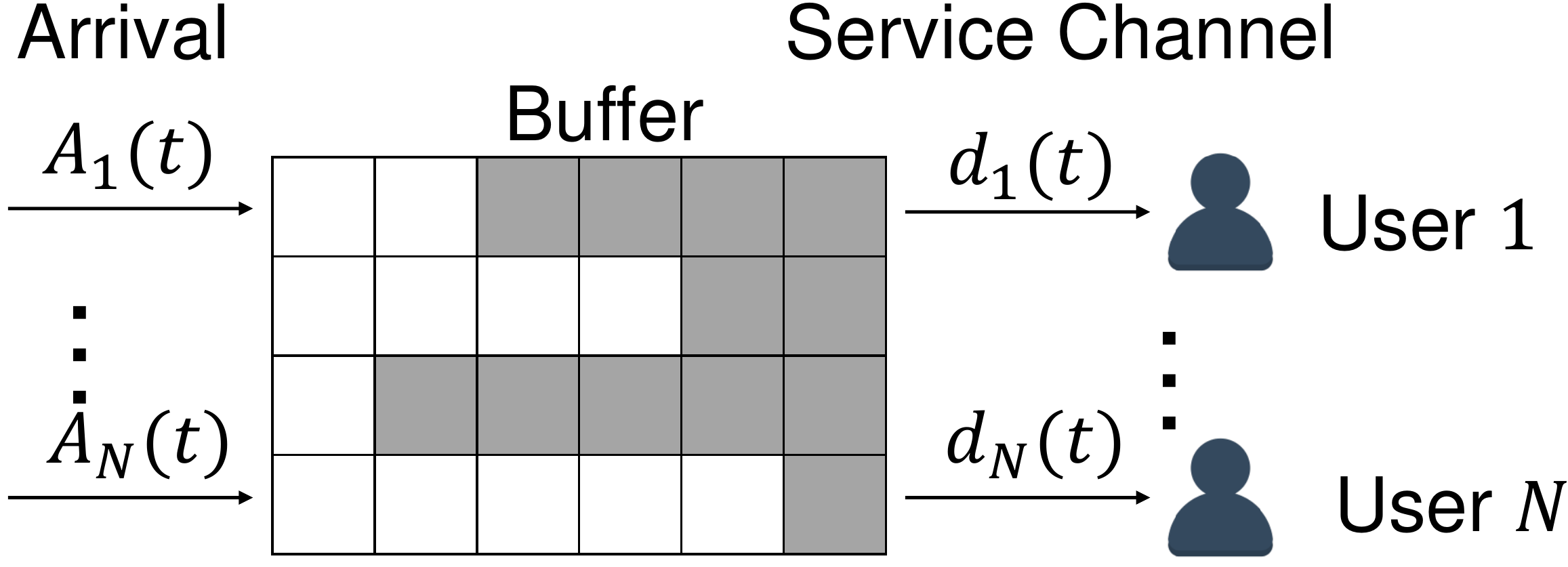}
		\caption{A general delay-constrained scheduling problem in a single-hop network. Jobs arrive at the server and need to be delivered to their destinations before deadline.}
		\label{fig:sm}
		\vspace{-.6cm}
	\end{figure}

	We consider the scheduling problem illustrated in Figure \ref{fig:sm}.
	Time is divided into discrete slots $t\in\{0, 1, 2, ...\}$.
	At the beginning of each time slot, the scheduler first receives job arrivals, e.g., data packets in a network or parcels in a delivery station.
	The number of job arrivals for user $i$ at time slot $t$ is denoted as $A_i(t)$.
	We denote $\boldsymbol{A}(t)=[A_1(t), ..., A_N(t)]$. Each job for user $i$ is associated with a strict delay constraint $\tau_i$, i.e., a job needs to be served within $\tau_i$ slots upon its arrival and will be outdated and discarded otherwise.
    
	\paragraph{The buffer model}
	Jobs arriving at the system are first placed in a buffer, conveniently modeled by a set of delay-sensitive queues.
	Specifically, the buffer contains $N$ separate delay-sensitive queues of infinite size, one for each user.
	The state of each queue $i$ at time slot $t$ is denoted by  $\boldsymbol{B}_i(t)=[B_i^{0}(t),B_i^{1}(t),...,B_i^{\tau_i}(t)]$,  where $B_i^{\tau}(t)$ stands for the number of jobs for user $i$ with a remaining time of $\tau$ timeslots until expiration for $1\le \tau\le\tau_i$.

	\paragraph{The scheduling and service model}
	At every time slot $t$, the scheduler makes decision on the resources allocated to jobs in the buffer. The decision for user $i$ is denoted as $\boldsymbol{e}_i(t)=[e_i^{0}(t),e_i^{1}(t),...,e_i^{\tau_i}(t)]$, where $e_i^{\tau}(t)\in[0, e_{\max}]$ denotes the resource, e.g., energy,  allocated to serve each job in queue $i$ with deadline $\tau$.
	Each scheduled job then goes through a service channel, e.g., a wireless channel,  whose  condition is random and its value at time slot $t$ for user $i$ is denoted by $c_i(t)$.
	We denote the service channel conditions at timeslot $t$ as $\boldsymbol{c}(t)=[c_1(t),c_2(t),...,c_N(t)]$.
	For a user $i$ job with allocated resource $e$ and channel condition $c$, the probability of successful service is denoted as $P_i(e,c)$, where $e=0$ means that the job will not be served and $P_i(0,c)=0$.
	Also, $P_i(e,\cdot)>0$ for all $e>0$.
	If a job is scheduled but fails in service, it remains in the buffer if it is not outdated.
	The instantaneous resource consumption at time slot $t$ is denoted as $E(t)=\sum_{i=1}^N\boldsymbol{e}_i^\top(t)\boldsymbol{B}_i(t)$ and the average resource consumption is $\overline{E}=\lim_{T\to\infty}\frac{1}{T}\sum_{t=1}^TE(t)$. 

    \paragraph{The system objective}
    For user $i$, the number of successfully served jobs at time slot $t$ is denoted as $d_i(t)$. Each user is given a weight $\beta_i$, and the weighted instantaneous throughput is denoted as $D(t)=\sum_{i=1}^N\beta_id_i(t)$.
    The objective of the scheduler is to maximize the weighted average throughput, defined as $\overline{D}=\lim_{T\rightarrow\infty}\frac{1}{T}\sum_{t=1}^TD(t)$, subject to the average resource consumption limit, i.e., \footnote{We assume w.l.o.g. that all corresponding limits exist. The results can be generalized with $\limsup$ and $\liminf$ definitions otherwise.
    }
	\begin{eqnarray}\label{eq:p1}
\mathcal{P}\quad	\max_{\boldsymbol{e}_i(t):1\le i\le N,1\le t\le T}&&\lim_{T\rightarrow\infty}\frac{1}{T}\sum_{t=1}^T\sum_{i=1}^N\beta_id_i(t)\\
		\text{s.t.}&&\lim_{T\to\infty}\frac{1}{T}\sum_{t=1}^T\sum_{i=1}^N\boldsymbol{e}_i^\top(t)\boldsymbol{B}_i(t)\le E_0\notag
	\end{eqnarray}
    where $E_0$ is the average resource consumption limit. We denote its optimal value as $\mathcal{T}^*$. 
    
    Note that although our problem description adapts the formulation in \cite{singh2018throughput}, our goal is to develop practical and scalable solutions to scheduling in real-world environments, i.e.,  without assumptions on ergodicity of dynamics and allow the randomness to be arbitrarily correlated with hidden factors. 
    Note here we allow resources to be allocated without hard constraint. We will show in Section \ref{subsec:pc} that hard constraints can be easily  incorporated with only minimal impact on performance. 
    
    \vspace{-.2cm}
    \subsection{Lagrange Dual}\label{subsec:ld}
    Define the following Lagrangian function of problem (\ref{eq:p1}) for handling the average resource budget $E_0$: 
	\begin{equation}\label{eq:lg}	
	\mathcal{L}(\pi,\lambda)=\lim_{T\rightarrow \infty}\frac{1}{T}\sum_{t=1}^T\sum_{i=1}^N[\beta_id_i(t)-\lambda\boldsymbol{e}_i^\top(t)\boldsymbol{B}_i^\top(t)]+\lambda E_0
	\end{equation}
	with respect to control policy $\pi$ and Lagrange multiplier $\lambda$.
	Denote $g(\lambda)$ the Lagrange dual function for fixed Lagrange multiplier $\lambda$:
	\begin{equation}
		g(\lambda)=\max_\pi\mathcal{L}(\pi,\lambda)=\mathcal{L}(\pi^*(\lambda),\lambda),
	\end{equation}
	where the maximizer is denoted as $\pi^*(\lambda)$.
	Using Lemma 3 in \cite{singh2018throughput}, the optimal timely throughput $\mathcal{T}^*$ equals the optimal value of the dual problem, i.e., $\mathcal{T}^*=\min_{\lambda\ge0}g(\lambda)=g(\lambda^*)$, where $\lambda^*$ is the optimal Lagrange multiplier. 
 To find the optimal policy $\pi^*(\lambda^*)=\arg\max_{\pi}\mathcal{L}(\pi,\lambda^*)$, it suffices to find optimal Lagrangian multiplier $\lambda^*$.
	Denote the consumed resource under policy $\pi$ in time slot $t$ as $E_\pi(t)$.
	The Danskin's Theorem in  \cite{bertsekas2003convex} states that 
	$	g'(\lambda)  = \frac{\partial\mathcal{L}(\pi^*, \lambda)}{\partial \lambda} = E_0 - E_{\pi^*(\lambda)}$, 
	where $E_{\pi^*(\lambda)}=\lim_{T\to\infty}\frac{1}{T}\sum_{t=1}^TE_{\pi^*(\lambda)}(t)$ denotes the average resource consumption.
	Thus, the optimal policy $\pi^*(\lambda^*)$ can be obtained by recovering the dual function $g(\lambda)$ for some $\lambda$ and taking gradient descent to find the optimal $\lambda^*$.
	While the theoretical understanding is clear, 
 finding the optimal $\pi^*(\lambda)$ for a given $\lambda$ in a practical system is non-trivial, due to the following main reasons: 
	\begin{itemize}
	    \item System dynamics are hard to trace since distributions can be highly dynamic and correlated in many scenarios.
	    \item The system can only be  partially observed, e.g., due to sensing limitation and noise and other hidden factors. 
	    \item Practical scheduling systems usually have a large number of users and complex multi-hop topology, which demand highly scalable solutions. 
	\end{itemize}
	Prior works \cite{singh2018throughput, chen2018timely} uses DP to find the maximizer $\pi^*(\lambda)$, which requires the prior knowledge of system dynamics and suffers from the curse-of-dimensionality in large systems, and may not directly apply to partially observable systems. 
	These motivate us to design a DRL-based framework with POMDP formulation in Section \ref{sec:framework}. 
	

    
    \vspace{-.2cm}
    \section{Overall Framework}\label{sec:framework}
    In this section, we present our novel POMDP formulation for solving the scheduling problem $\mathcal{P}$ in Section \ref{subsec:sm},
    which supports  user-level decomposition and node-level merging for scalability.

	\vspace{-.2cm}
	\subsection{The POMDP Formulation}

    We now specify the POMDP formulation for the scheduling problem, under which the $\mathtt{RSD4}$ can be applied
    to find the optimal policy $\pi^*(\lambda^*)$.
	Specifically, the POMDP is represented by $\mathcal{M}=\langle\mathcal{S},\mathcal{O},\mathcal{A},r,P,\gamma\rangle$, where $\mathcal{S}$ is the state space, $\mathcal{O}$ is the observation space, $\mathcal{A}$ is the action space, $r$ is the reward function, $P$ denotes the transition matrix, and $\gamma$ stands for the discount factor.
	
\begin{figure}[htbp]
		\centering
		\vspace{-.4cm}
		\subfigure[Observed system\label{subfig:observation}]{
			\includegraphics[width=.45\columnwidth]{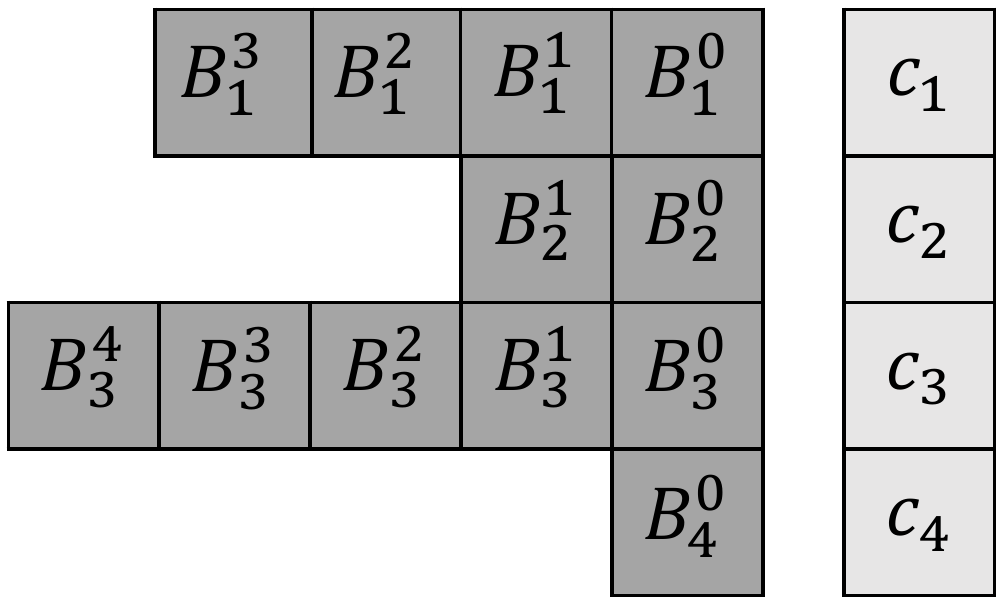}
		}
		\subfigure[Decomposed observation\label{subfig:decomp_ob}]{
			\includegraphics[width=.45\columnwidth]{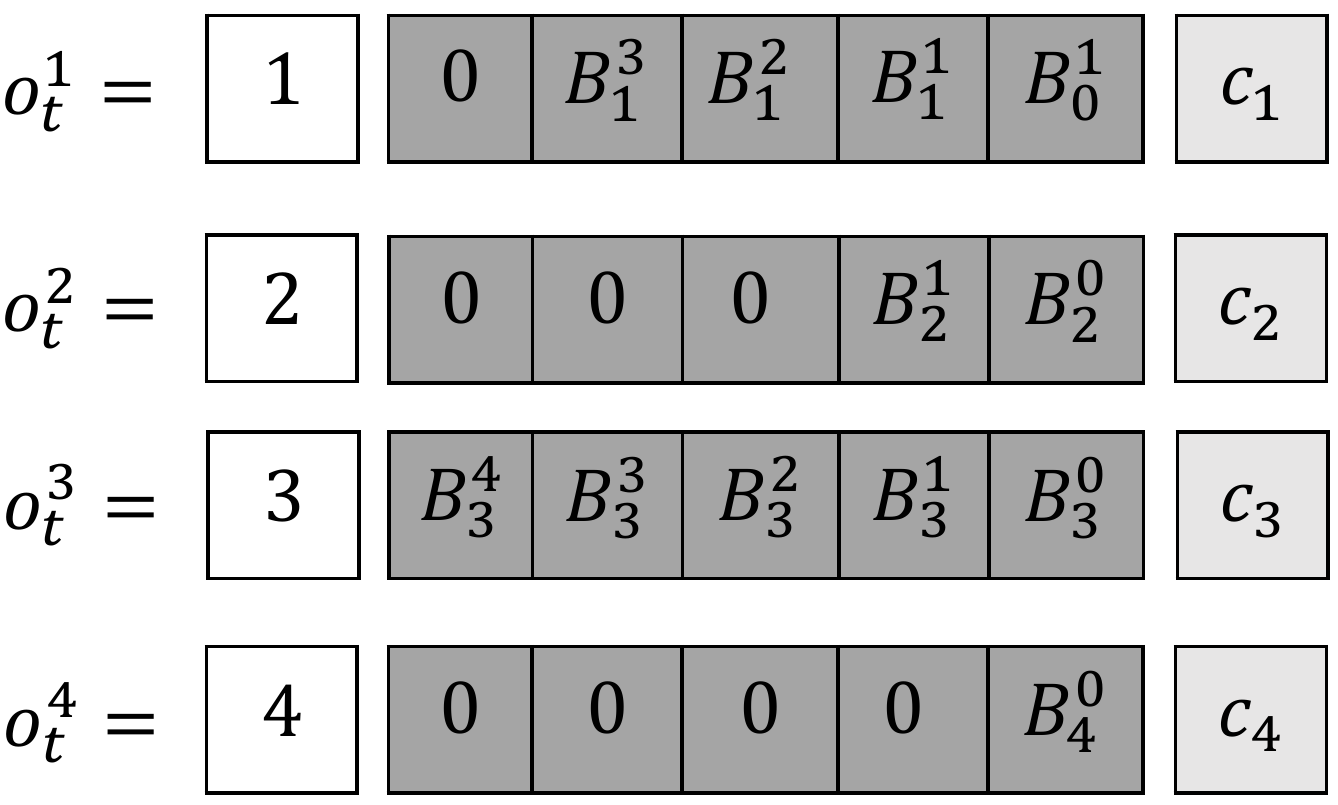}
		}
		\subfigure[System observation\label{subfig:full_ob}]{
			\includegraphics[width=.9\columnwidth]{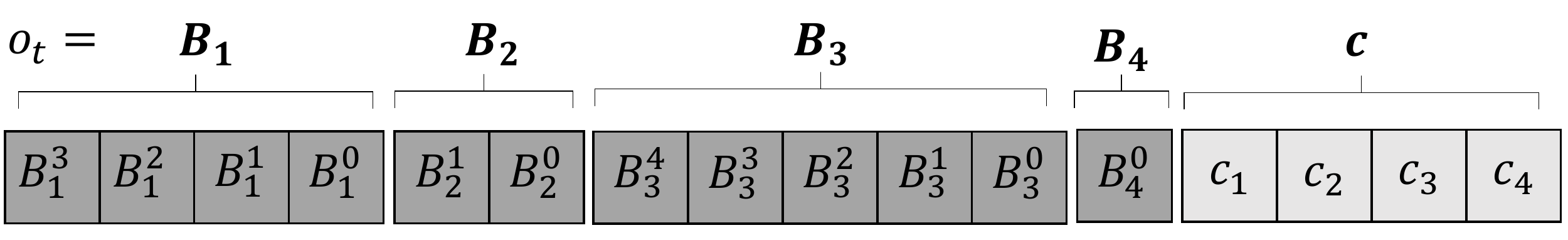}
		}
		\vspace{-.6cm}
		\caption{Observations for a four-user system (time index omitted for brevity).}
		\label{fig:tmp}
		\vspace{-.6cm}
	\end{figure}

	\paragraph{State and Observation}
	The overall system state $s_t$ includes $\boldsymbol{A}(t)$, $\boldsymbol{B}_1(t)$,..., $\boldsymbol{B}_N(t)$, $\boldsymbol{c}(t)$ and other   information of the underlying MDP unobservable by the scheduler, e.g., random hyperparameters of successful transmission probability, or mobile user position that affects traffic arrival and channel. 
	We consider $s_t$ to be partially-observed since partial observable scenarios are common in scheduling problems.
	This implies that the actual observation $o_t$ is a subset of $s_t$ and the exact form of $o_t$ depends on environment settings.
	For example, Figure \ref{subfig:observation} shows the observed system of a four-use case, where the observation $o_t=[\boldsymbol{B}_1(t),...,\boldsymbol{B}_N(t),\boldsymbol{c}(t)]$ only includes buffer with delay-sensitive queues  and channel states but not other factors. 
	
	\paragraph{Action and Reward}
	At time slot $t$, the control action is denoted by  $a_t=[\boldsymbol{e}_1(t),...,\boldsymbol{e}_N(t)]$, and the reward is set to 
	\begin{eqnarray}
		r_t=D(t)-\lambda E(t), \label{eq:reward}
	\end{eqnarray}
	which is the instantaneous weighted throughput $D(t)$ minus the resource consumption $E(t)$ weighted by the multiplier $\lambda$.
	
	\paragraph{Learning Objective}
	Under POMDP, an optimal agent needs to access the entire history $h_t = (o_1,a_1,o_2,a_2,...,a_{t-1}, o_t)$ and learn a deterministic policy $\pi(\cdot; \phi)$ parameterized by $\phi$, which maps from the history to the action space, with the objective of maximizing the expected long-term rewards
	$$J(\pi(\cdot; \phi))=\mathbb{E}\left[\sum_{t=0}^{T} \gamma^{t} r\left(s_{t}, a_{t}\right) \mid s_{0}, a_{0}, \pi(\cdot; \phi)\right].$$
	
	\vspace{-.2cm}
	\begin{remark}
		With the reward setting in (\ref{eq:reward}), when $\gamma=1$, the cumulative discounted reward is  $R=\sum_{t=1}^T\gamma^tr_t=T{\mathcal{L}}(\pi,\lambda)-\lambda TE_0$, which differs from the Lagrangian function value in (\ref{eq:lg}) by $-\lambda TE_0$. 
		This means that an algorithm maximizing the expected rewards $J(\pi(\cdot;\phi))$ is also a maximizer for the Lagrangian function, which is the objective of $\mathtt{RSD4}$ (detailed in Section \ref{sec:rsd4}).
	\end{remark}

	
	\vspace{-.3cm}
	\subsection{User-Level Decomposition}\label{subsec:ds}
	The system state contains the buffer information whose size is proportional to the user number. This makes learning much harder in large-scale systems, due to the need to train neural networks with more hyperparameters, which largely limits the scalability of DRL-based methods.
	%
	To overcome this issue, we propose a \emph{user-level decomposition} technique, such that $\mathtt{RSD4}$ finds the optimal policy with small neural networks even under large-scale scenarios. 
	Our user-level decomposition is  different from the packet-level decomposition in \cite{singh2018throughput, chen2018timely}, primarily due to the fact that packet-level decomposition requires a much larger number of neural network parameters, resulting in poor performance. 
	
	
	Specifically, we define the user-level decomposed Lagrangian function for user $i$ as
	\begin{eqnarray}\label{eq:lgri}	{\mathcal{L}}_i(\pi_i,\lambda_i)=\lim_{T\to\infty}\frac{1}{T}\Big[\sum_{t=1}^T\beta_id_i(t)-\lambda_i\boldsymbol{e}_i^\top(t)\boldsymbol{B}_i^\top(t)\Big],
	\end{eqnarray}
	where $\pi_i$ is the decomposed policy for scheduling user $i$'s jobs in the buffer.
	Consequently, maximizing the Lagrangian function in (\ref{eq:lg}) for a fixed $\lambda$ can be accomplished by maximizing (\ref{eq:lgri}) for each user $i$'s sub-problem separately with $\lambda_i=\lambda$.
	Denote the optimal policy for user $i$'s sub-problem as ${\pi}_i^*(\lambda_i)$ with multiplier $\lambda_i$.
	After solving Problem (\ref{eq:lgri}) for each user with a fixed $\lambda_i$, the optimal policy ${\pi}^*(\lambda)$ that maximizes the Lagrangian (\ref{eq:lg}) can be derived by letting each user take its own optimal scheduling policy ${\pi}_i^*(\lambda_i)$ with $\lambda_i=\lambda$.

	\paragraph{Decomposed POMDP}
	Based on the above intuition, we decompose the POMDP into user-level subproblems, where for each user $i$, 
	the action is  $a_t^{(i)}=\boldsymbol{e}_i(t)$, and the reward becomes $r_t^{(i)}=\beta_id_i(t)-\lambda\boldsymbol{e}_i^\top(t)\boldsymbol{B}_i(t)$ at time slot $t$.
	The observation is also decomposed as well, Figure \ref{subfig:decomp_ob} presents  $o_t^{(i)}=[i,\boldsymbol{B}_{i}(t),\boldsymbol{c}_i(t)]$ for a four-user case, where the first index for user $i$'s sub-problem is required for algorithm training (explained next). This is a key step in $\mathtt{RSD4}$. As	we will see in Section \ref{subsec:scalability}, without decomposition, $\mathtt{RSD4}$ and other existing DRL algorithms can fail due to a large size of the state representation. 
	
	
	\paragraph{Unified Training}
	After decomposition, the state space is compressed, and there are $N$ different sub-POMDPs, whose dynamics of different users may not be the same.
	To avoid training $N$ different DRL agents, leading to linear growth of computation power, we propose the method of unified training.
	That is, to train different samples from different sub-environments together, and use an extra user index $i$ as the identifier to samples from the user $i$'s sub-problem, as shown in Figure \ref{subfig:decomp_ob}.
	Consequently, the dimension of training samples remains the same regardless of the system scale.
	
	\begin{remark}
	    With observation decomposition and unified training, a single system observation $o_t$ is decomposed into $N$ separate sate $o_t^1,o_t^2,...,o_t^N$, such that one step dynamics in the original environment creates $N$ samples for the replay buffer.
	    Consequently, it does not require heavy parallel computation and greatly enriches the abundance of the replay buffer, such that common knowledge across different users' sub-POMDPs is learned efficiently.
	    With PODMP decomposition, the number of neural network parameters also remains small even for large-scale systems, because the dimension of observation remains unchanged after decomposition. This useful feature makes our framework highly scalable. 
	\end{remark}
	
	\subsection{Multi-Hop and Node-level Merging}\label{subsec:multihop}
\begin{figure}[htbp]
	    \vspace{-.5cm}
		\centering
		\subfigure[Multi-hop network with three paths.\label{subfig:network}]{
			\includegraphics[width=.45\columnwidth]{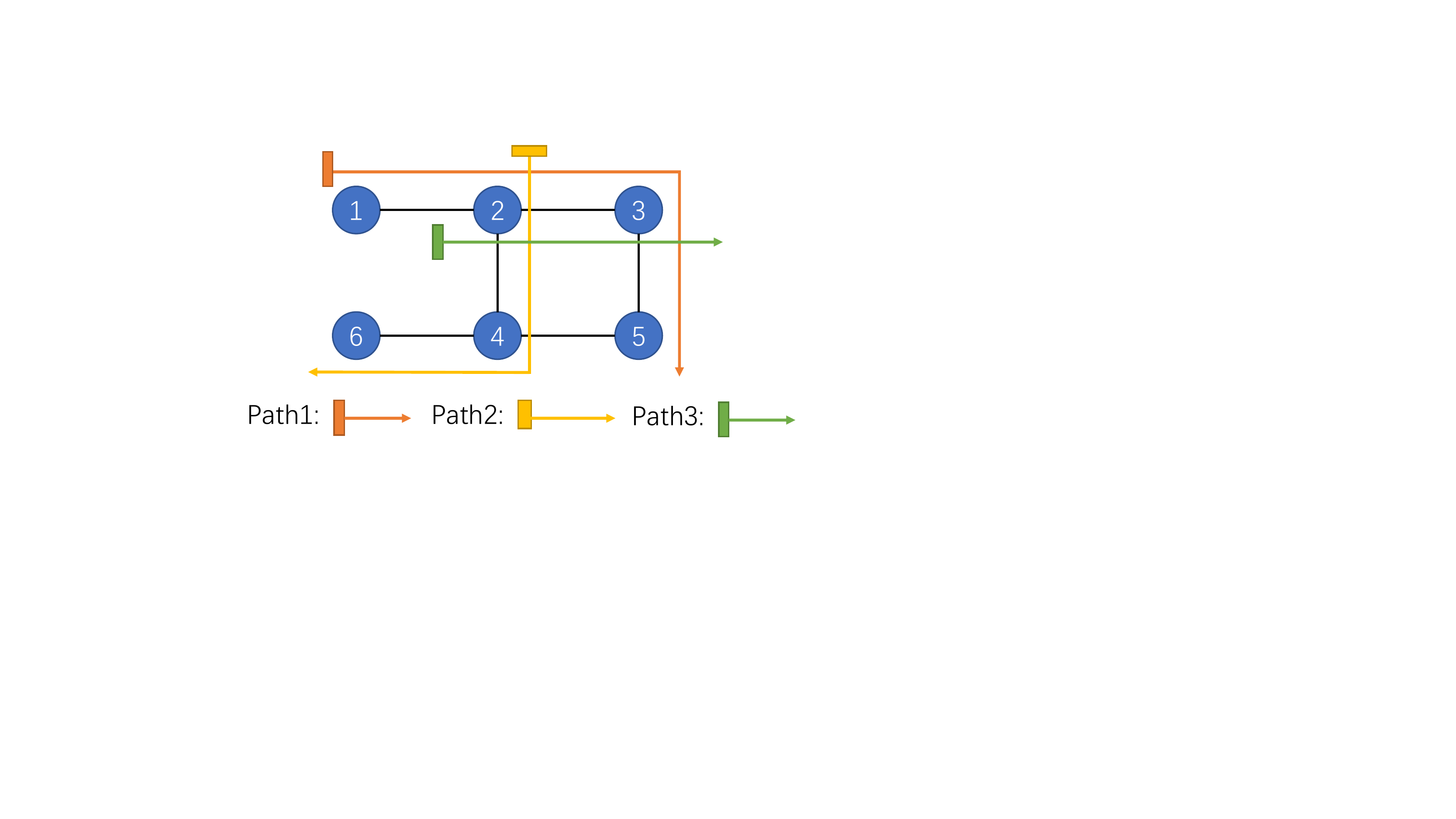}
		}
		\subfigure[Three flows\label{subfig:flow}]{
			\includegraphics[width=.4\columnwidth]{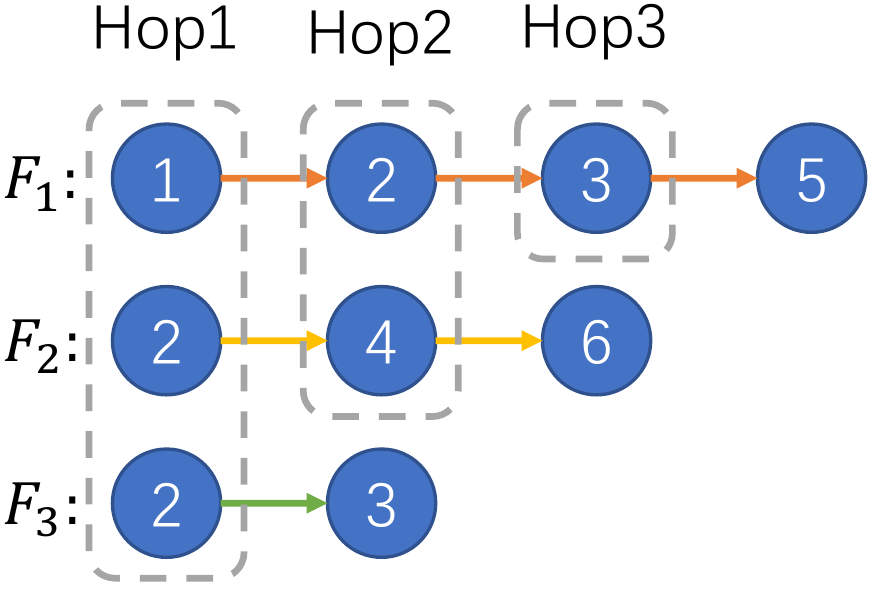}
		}
		\vspace{-.4cm}
		\caption{(a) A multi-hop network with three paths represented in different colors, each supports one flow. (b) The flows are aligned with starting nodes.}
		\label{fig:network}
		\vspace{-.3cm}
	\end{figure}

\begin{figure}[htbp]
	 	\vspace{-.2cm}
    	\centering
    	\includegraphics[width=.9\columnwidth]{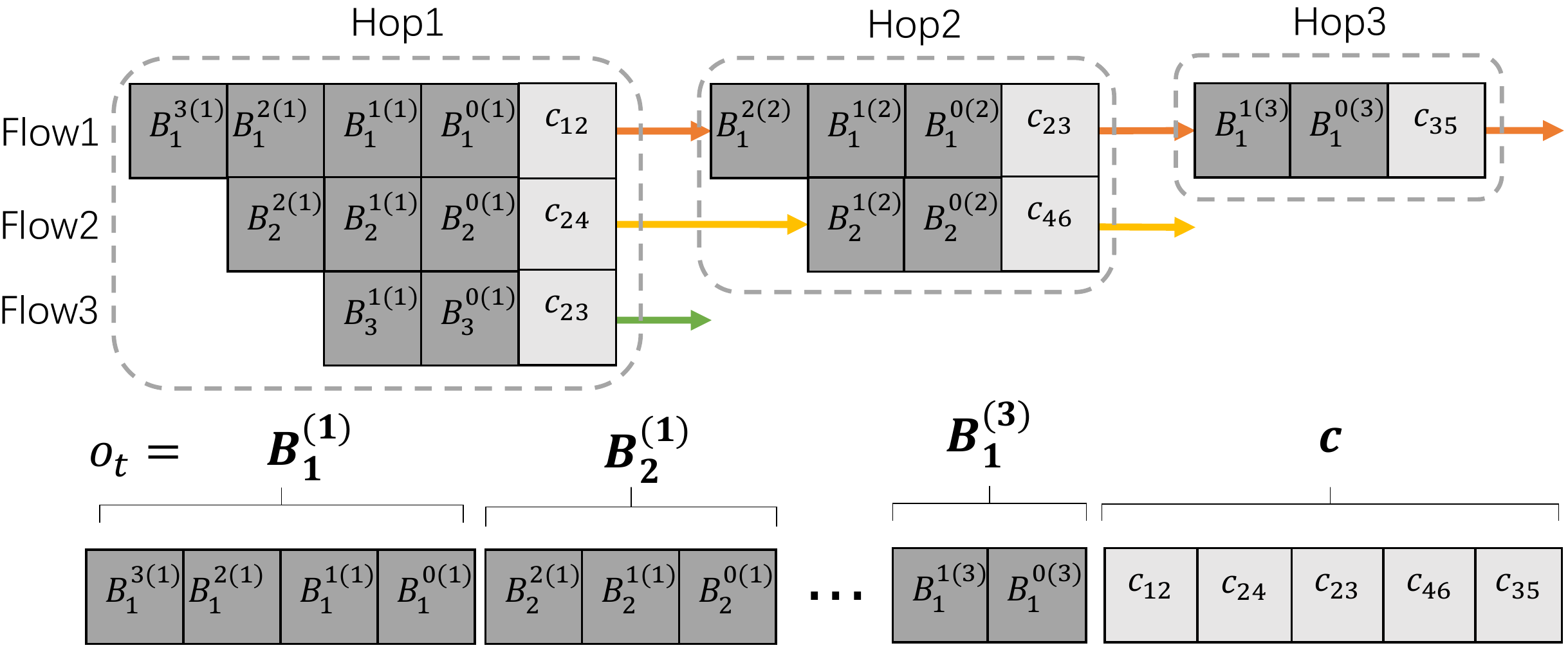}
    	\vspace{-.2cm}
    	\caption{Node-level Merging of a three-flow case. Flow $i$' node $j$ is associated with a buffer $\boldsymbol{B}_j^{i}$. Flows are aligned with starting nodes then concatenated as observation $o_t$.
    	}
    	\label{fig:merging}
    	\vspace{-.4cm}
    \end{figure}

	We present the multi-hop setting here and describe a technique called \emph{node-level merging} for reducing complexity in this case.
	Specifically, in multi-hop networks, each flow can traverse multiple hops and the agent needs to decide which flows to serve and how many resources to allocate at each node.
	Besides, each node has an average resource constraint.
	Take Figure \ref{subfig:network} as an example. There are three flows passing through three paths respectively, i.e.,   $F_1=\{1\rightarrow2\rightarrow3\rightarrow5\}$ with deadline $\tau_1$, $F_2=\{2\rightarrow4\rightarrow6\}$ with deadline $\tau_2$, and $F_3=\{2\rightarrow3\}$ with deadline $\tau_3$.
	The idea of node-level merging is to augment the state, observation, action, and reward in the POMDP formulation for multihop scheduling. Figure \ref{fig:merging} shows one example with the detailed procedure below. 
	
	\begin{itemize}
	    \item \emph{State and Observation}: The system state is obtained by merging buffer and channel states in different hops, as shown in Figure \ref{fig:merging}.
	    Since jobs in a multi-hop flow differ only in node position and remaining time until expiration, we encode the system state by an aggregation of buffer states at the nodes. 
	    Besides, the buffer state at hop $j$ for flow $i$ is denoted as $\boldsymbol{B}_i^{(j)}(t)=[B_i^{0(j)}(t),B_i^{2(j)}(t),...,B_{i}^{\tau_i(j)}(t)]$ where $B_i^{\tau(j)}(t)$ denotes the number of jobs for flow $i$ at hop $j$ with a remaining time of $\tau$ timeslots until expiration for $0\le\tau\le\tau_i$.
	    Denote the path length flow $i$ as $h_i$.
	    The node-level merging idea is to set the overall observation  $o_t$ to be  $[\boldsymbol{B}_1^{(1)}(t),...,\boldsymbol{B}_1^{(h_1)}(t),...,\boldsymbol{B}_N^{(h_N)}(t),\boldsymbol{c}(t)]$ and potentially other factors in the environment.  Figure \ref{fig:merging} shows a case when $o_t=[\boldsymbol{B}_1^{(1)}(t),...,\boldsymbol{B}_1^{(h_1)}(t),...,\boldsymbol{B}_N^{(h_N)}(t),\boldsymbol{c}(t)]$. 
	    
	    \item \emph{Action}: The action is similarly obtained by merging actions at different nodes, i.e.,  $a_t=[\boldsymbol{e}_1^{(1)}(t),...,\boldsymbol{e}_1^{(h_1)}(t),...,\boldsymbol{e}_N^{(h_N)}(t)]$, where $\boldsymbol{e}_i^{(j)}(t)$ represent the resource allocated to user $i$'s $j$-th node for $1\le j\le h_i$.
	    
	    
	    \item \emph{Reward}: The reward is set to $r_t=D(t)-\sum_{k=1}^M\lambda_kE^{(k)}(t)$, where $\lambda_k$ and $E^{(k)}(t)$ correspond to the Lagrangian multiplier and resource consumption at node $i$ for $1\le k\le M$, and $M$ is the number of nodes involved in the scheduling. 
	\end{itemize}
    A common approach for multi-hop scheduling is to address the scheduling problem in each node separately.
    However, in delay-constrained multi-hop networks, scheduling each node separately cannot satisfy the hard delay constraint.
    Multi-agent architecture \cite{xu2020joint, zhou2021multi} has been adopted for multi-hop networks but with more computation resource requirements and is hard to train.
    Node-level merging provides a novel way of concatenating buffers of different hops in flows together, making its training  similar to that for a single-hop scheduling problem except with a higher input dimension. This  avoids training multiple agents for multi-hop networks, and  efficiently reduces the complexity in training. It allows $\mathtt{RSD4}$ to perform well and outperforms other methods, even when the system scale in terms of state dimension is significantly increased in multihop networks (see Section \ref{subsec:scalability}).


	
    \section{Proposed Method: $\mathtt{RSD4}$}\label{sec:rsd4}
	We present our novel \underline{R}ecurrent \underline{S}oftmax \underline{D}elayed \underline{D}eep \underline{D}ouble \underline{D}eterministic Policy Gradient ($\mathtt{RSD4}$) algorithm in Algorithm \ref{alg:rsd4}.
	$\mathtt{RSD4}$ builds upon the recurrent deterministic policy gradient \cite{heess2015memory} and softmax deterministic policy gradient \cite{pan2020softmax}, and introduces several novel components for handling the scheduling problem in partially observed settings. 
	Specifically, $\mathtt{RSD4}$ is a model-free DRL algorithm, which takes the popular actor-critic \cite{konda2000actor} framework.
	It well handles partial observability issues by memory mechanism enabled by recurrent neural networks (RNNs) and resolves the overestimation problem in existing recurrent DRL methods with the softmax operator. Thus, it owns advantages from recurrent deterministic policy gradient \cite{heess2015memory} and softmax deterministic policy gradient \cite{pan2020softmax}.
	$\mathtt{RSD4}$ also adopts a double-branch architecture from \cite{peng2018sim} to better utilize the memory mechanism of the RNN, and implements a delayed policy update frequency to further reduce the variance in value estimate. 
	
	Initially, $\mathtt{RSD4}$ makes use of the state-action function $Q(h_t,a_t;\theta)$ parameterized by $\theta$, which is defined as
	\begin{equation}
		\resizebox{\columnwidth}{!}{$Q(h_t,a_t;\theta)=\mathbb{E}_{s_t,a_t,...,s_{t+T},a_{t+T}|h_t,\pi(\cdot;\phi)}\Big[\sum_{i=0}^T\gamma^ir(s_{t+i},a_{t+i})\Big]$},
	\end{equation}
	where the expectation is taken with respect to the conditional probability  $p(s_t,a_t,...,s_{t+T},a_{t+T}|h_t,\pi(\cdot;\phi))$ of the trajectory distribution associated with history $h_t$ and the policy $\pi(\cdot;\phi))$.
	$\mathtt{RSD4}$ initializes double critic networks and double actor networks, where critic networks $Q(h_t,a_t;\theta)$   estimate the value of state-action pairs, and actor networks $\pi(\cdot;\phi)$ are responsible for outputting control actions.

        \begin{algorithm}[htbp]
		\caption{$\mathtt{RSD4}$ with decomposition}
		\label{alg:rsd4}
		\begin{algorithmic}[1]
			\REQUIRE $\lambda_0$, resource limit $E_0$, learning rate $\alpha$, Precision $\delta$, episode number $M$, episode length $T$, batch size $b$, target update rate $\tau$.
			\STATE $\lambda_1\leftarrow\lambda_0, \lambda_0\leftarrow0, k\leftarrow1$
			\WHILE {$|\lambda_{k}-\lambda_{k-1}|>\delta$}
			\STATE Initialize $N$ learning environments with $r_t^{(i)}=\beta_id_i(t)-\lambda\boldsymbol{e}_i^\top(t)\boldsymbol{B}_i(t)$ for $i$-th environment.\hspace{-.3cm}
			\STATE Initialize critic networks $Q_1$, $Q_2$, and actor networks $\pi_1$, $\pi_2$ with random parameters $\theta_1,\theta_2, \phi_1, \phi_2$. Initialize target network $\theta_1^-\leftarrow\theta_1,\theta_2^-\leftarrow\theta_2,\phi_1^-\leftarrow\phi_1,\phi_2^-\leftarrow\phi_2$.
			\STATE Initialize replay buffer $\mathcal{D}$.
			\FOR{episodes $=1$ to $M$\textcolor{blue}{ // Episodic interaction}}
			\FOR{$t=1$ to $T$}
			\FOR{$i=1$ to $N$}
			\STATE Receive sub observation $o_t^i$
			\STATE $h_t^i\leftarrow h_{t-1}^i,a_{t-1}^i,o_t^i$
			\STATE Select action $a_t^i$ based on $\pi_1$ and $\pi_2$.
			\ENDFOR
			\ENDFOR
			\STATE Store  $(o_1^i,a_1^i,r_1^i,...,o_T^i,a_T^i,r_T^i)$ in $\mathcal{D}$ for $i=1$ to $N$.
			\hspace{-.3cm}
			\FOR {$i=1,2$ \textcolor{blue}{// Double learning}}
			\STATE Randomly sample a batch of $b$ episodes: $\mathcal{B}=\{(o_1,a_1,r_1,...,o_T,a_T,r_T)\}$ from $\mathcal{D}$.
			\FOR{$t=1$ to $T$ \textcolor{blue}{// Recurrent softmax learning}}
			\STATE Sample $K$ noise $\epsilon\sim\mathcal{N}(0,\sigma')$
			\STATE $\hat{a}_t\leftarrow\pi_i(h_t;\phi_i^-)+clip(\epsilon,-c,c)$
			\STATE $\hat{Q}(h_t, \hat{a}_t) \leftarrow \min _{j=1,2}(Q_{j}(h_t, \hat{a}_t; \theta_{j}^{-}))$
			\STATE Compute $\operatorname{softmax}_{\beta}(\hat{Q}(h_t, \cdot))$ by Eq. (\ref{eq:sm})
			\STATE $y_{t} \leftarrow r+\gamma(1-d) \operatorname{softmax}_{\beta}(\hat{Q}(h_t, \cdot))$ 
			\ENDFOR
			\STATE Update $\theta_i$ according to Bellman loss:\\
			$\frac{1}{N} \sum_{h\in\mathcal{B}}\sum_t\left(Q_{i}\left(h_t, a ; \theta_{i}\right)-y_{i}\right)^{2}$
			\IF {episodes mod $d=0$ \textcolor{blue}{// Delayed update} } 
			\STATE Update actor $\phi_i$ by recurrent policy gradient:
			\begin{equation}
			    \vspace{-.3cm}
				\begin{split}
					\nabla_{\phi_i} J(\phi_i)=&\frac{1}{N}\sum_{h\in\mathcal{B}} \sum_{t}\Big[\nabla_{\phi_{i}}\left(\pi(h_t ; \phi_{i})\right) \\
					& \nabla_{a} Q_{i}(h_t, a ; \theta_{i})|_{a=\pi\left(h_t ; \phi_{i}\right)}\Big]
				\end{split}
				\nonumber
				\vspace{-.3cm}
			\end{equation}
			\STATE Update target networks:\\
			$\theta_{i}^{-} \leftarrow \tau \theta_{i}+(1-\tau) \theta_{i}^{-}$, $\phi_{i}^{-} \leftarrow \tau \phi_{i}+(1-\tau) \phi_{i}^{-}$
			\ENDIF
			\ENDFOR
			\ENDFOR
			\STATE Obtain policy $\pi_k$ and evaluate $E_{\pi_k}.$
			\STATE $\lambda_{k}=\lambda_{k}+\alpha(E_{\pi_k}-E_0)$
			\textcolor{blue}{// Gradient update}
			\STATE $\lambda_{k-1}\leftarrow\lambda_k$
			\ENDWHILE
			\STATE Output $\pi_k$
		\end{algorithmic}
	\end{algorithm}
 
	\subsection{The Training Algorithm}
	\paragraph{Recurrent Network Architecture}
	Despite the success of RL in a number of challenging tasks, state-of-the-art RL algorithms such as TD3 \cite{fujimoto2018addressing} are limited to solving fully-observable tasks.
	As a result, they can fail when faced with partially observable tasks as the problem considered here. 
	To address this problem, compared to prior non-recurrent policy gradient methods, $\mathtt{RSD4}$ incorporates recurrency in designing the architecture for neural networks rather than simply using feedforward networks in the policy update. This strengthens the memory capability of $\mathtt{RSD4}$, and enables it to learn hidden factors or temporal correlation of the  system dynamics.
	We then update the policy by RDPG  \cite{heess2015memory}:
	\begin{equation}\label{eq:rg}
	    \nabla_{\phi} J(\pi(\cdot ; \phi))
	=\mathbb{E}_{h_t}\Big[\sum_t\nabla_{\phi}(\pi(h_t ; \phi)) \nabla_{a} Q(h_t, a ; \theta)\Big|{a=\pi(h_t ; \phi)}\Big].
	\end{equation}
    Here to compute the RDPG, a sequence of   episodes will be stored in the replay buffer $\mathcal{D}$ as a training sample (line $14$).
    $\mathtt{RSD4}$ computes target values $(y_1,y_2,....,y_T)$ for each sampled episode using the recurrent networks in lines $17-23$.
    Critics and actors are updated recurrently, as shown in line $24$ and $26$, respectively.
	
	\paragraph{Softmax Double Learning}
	Having a good estimate of the value function is critical for RL agents to achieve good performance \cite{pan2020softmax}.
	We therefore propose to incorporate the softmax operator for more accurate value function estimation  in different scenarios.
	In lines $21-22$, we compute the Q-function by softmax operator by:
	\begin{equation}\label{eq:sm}
		\operatorname{softmax}_{\beta}(Q(h, \cdot))=\frac{\mathbb{E}_{a \sim p}\left[\frac{\exp \left(\beta Q\left(h, a ; \theta\right)\right) Q\left(h, a ; \theta\right)}{p\left(a\right)}\right]}{\mathbb{E}_{a \sim p}\left[\frac{\exp \left(\beta Q\left(h, a ; \theta\right)\right)}{p\left(a\right)}\right]}
	\end{equation}
	where $\beta$ denotes the inverse temperature and $p$ is the sampling distribution.
	The target value for critic $Q_i$ is given by $y_{i}=r+\gamma \operatorname{softmax}_{\beta}(\hat{Q}(h,\cdot))$ in line 22, where $\hat{Q}(h,\cdot)$ denotes the value estimation function in line 20.
	
	$\mathtt{RSD4}$ further adopts a delayed policy update mechanism from \cite{fujimoto2018addressing} (line $25$) to avoid training divergence due to frequent updates of the policy. Thus, the policy network is updated at a lower frequency than the value network to minimize error before introducing a policy update, with the similar goal of making the scheduling policy more robust in various system dynamics.

        \vspace{-.3cm}
	\subsection{Network Architecture}
\begin{figure}[htbp]
		\centering
		\vspace{-.3cm}
		\includegraphics[width=.8\columnwidth]{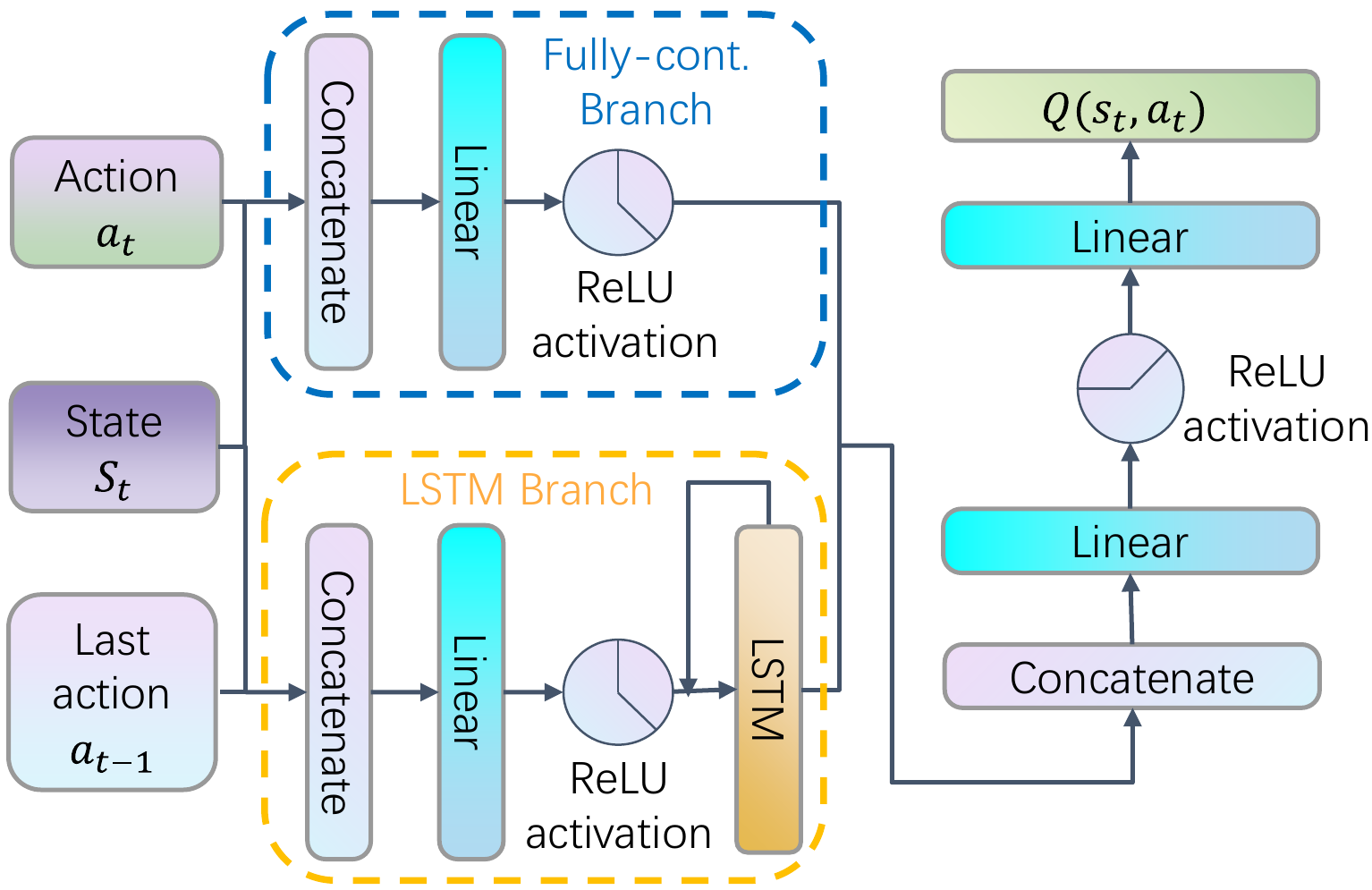}
		\vspace{-.3cm}
		\caption{The architecture of the critic network.
		There are double parallel branches of the LSTM branch and the fully-connected branch, which are later concatenated together by a fully-connected layer to output Q value.
		}
		\label{fig:critic}
		\vspace{-.6cm}
	\end{figure}

	$\mathtt{RSD4}$ uses double actor-networks and double critic-networks. The architecture of critic networks in Figure \ref{fig:critic}, and  actor networks are similar, with the difference being  removing the action $a_t$ in the input and changing the output value $Q(s_t,a_t)$ to action $a_t$. 
	The recurrent layers build upon Long-Short-Term-Memory (LSTM)  \cite{hochreiter1997long} to perform RDPG in Eq. (\ref{eq:rg}), and there are two parallel branches, i.e., the fully connected branch and the LSTM branch. This architecture is firstly proposed in \cite{peng2018sim} and is effective for our scheduling problem, as validated in our experiments. 
	
	The LSTM branch is designed to strengthen the memory ability of $\mathtt{RSD4}$ algorithm,  since it allows the agent to incorporate a large amount of network measurement history into its state space to capture the long-term temporal dependencies of actual system dynamics. The LSTM layer is embedded in the second layer of the multilayer perceptron feature extractor. 
	Consequently, our $\mathtt{RSD4}$ algorithm can well handle various partially observable settings. 
	The fully connected branch is designed to capture more information and improve expressiveness in the current time slot, which provides subsequent layers with more direct access to the current state without requiring information to filter through the LSTM branch. 
	This makes $\mathtt{RSD4}$ more sensitive to the current system state, so that abrupt changes of environmental conditions, e.g., a sudden burst  of arrivals or temporal channel condition degradation, can be detected rapidly (shown in Section \ref{subsubsec:sw}).
	
	\begin{remark}
	    The architecture in Figure \ref{fig:critic} accounts for the ability to resolve partial observability issues.
	    Most non-DRL-based methods and non-recurrent DRL algorithms do not directly handle this case.
	    Our POMDP formulation and $\mathtt{RSD4}$ solution well handle such potentially non-stationary and partially observable dynamics by the memory mechanism enabled by LSTM, which helps the agent to learn from a batch of history and reduce the impact of temporal variability.
	\end{remark}
	
	\section{Experiment Results}\label{sec:er}
	
	We conduct extensive experiments on   $\mathtt{RSD4}$ based on real-world datasets and simulated data. 
	We first present an ablation study of $\mathtt{RSD4}$ and then compare performances of $\mathtt{RSD4}$ with existing DRL methods and classical non-DRL-based algorithms. 
	Then, we design various hard settings to validate the ability of $\mathtt{RSD4}$ to handle partially observability and scalability.
	Each experiment is repeated with $5$ different random seeds, and the average result is presented.
 
	\subsection{Environment Setup}
    Below, we specify the single-hop environment based on the network in Figure \ref{fig:sm}.
	The multi-hop environment is constructed in a similar way and will be specified in Section \ref{subsec:scalability}.
	
\begin{figure}[H]
		\centering
		\vspace{-.4cm}
		\subfigure[Visulization of four arrivals in 5000 time slots.\label{subfig:hmarrival}]{
			\includegraphics[width=\columnwidth]{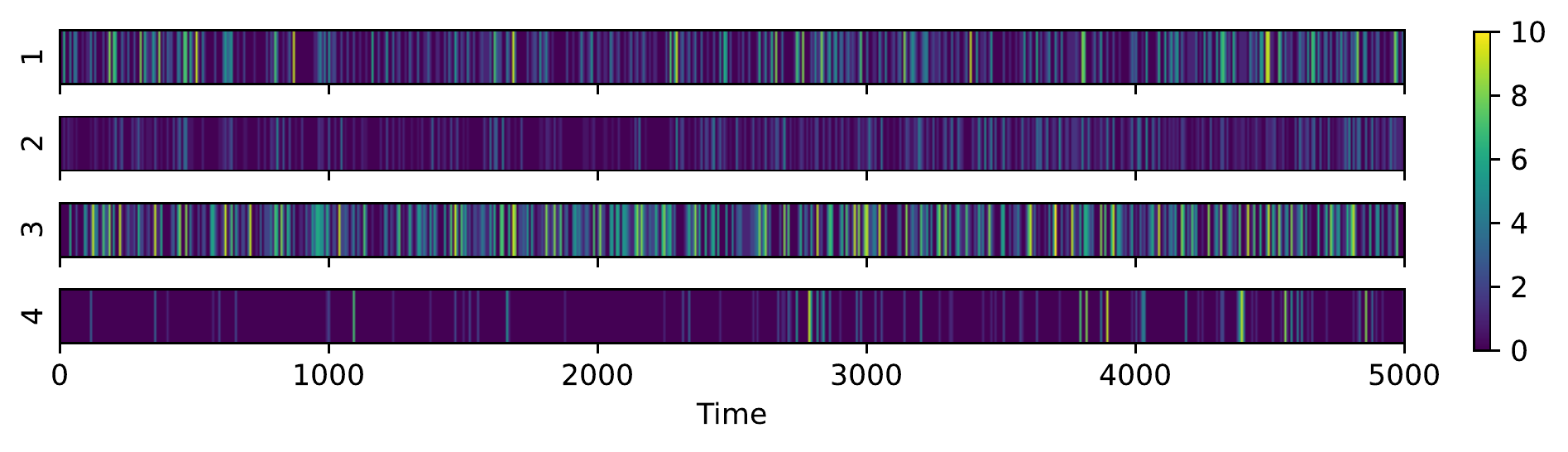}
		}
		\vspace{-.5cm}
		\subfigure[Visulization of channel conditions in 5000 time slots.\label{subfig:hmchannel}]{
			\includegraphics[width=\columnwidth]{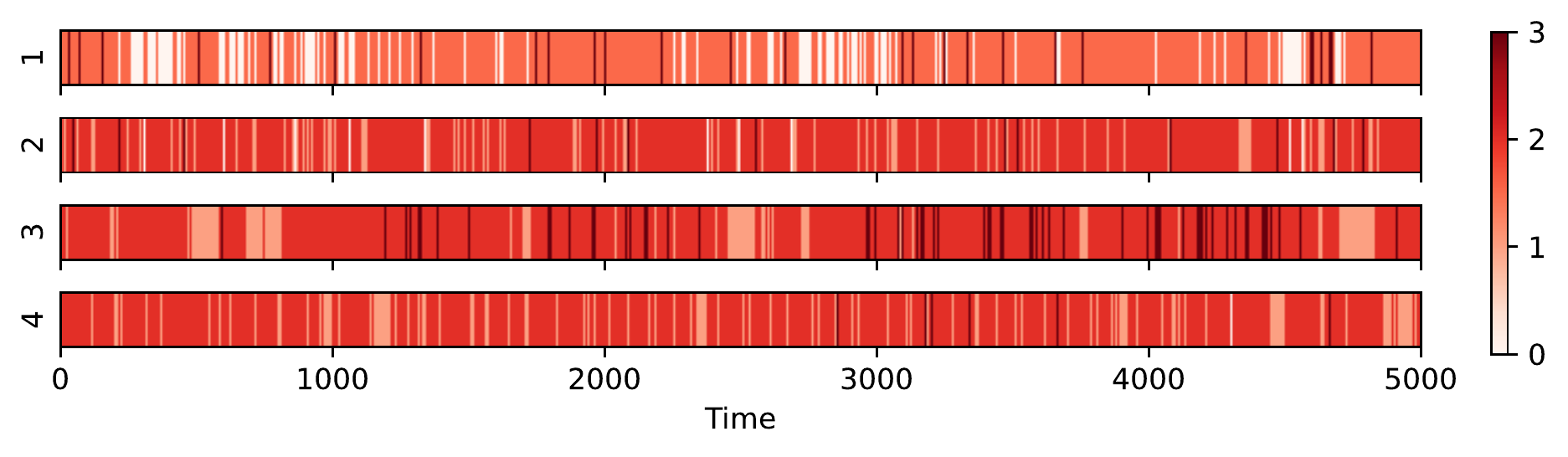}
		}
		\caption{Heat maps of arrivals and channel conditions.}
		\label{fig:heatmap}
		\vspace{-.5cm}
	\end{figure}

	\paragraph{Arrivals}
	The arrivals of different users are given by a LTE dataset \cite{ltedataset}, which records the traffic flow of mobile carriers' 4G LTE network in approximately one year.
	We construct an environment with four types of arrivals given by the LTE dataset, which are visualized in Figure  \ref{subfig:hmarrival}.
	The character of selected data records is given in Table \ref{tab:info}, which simulates four representative tasks, i.e., file transmission, online forum, VR gaming, and text communication, according to their rates and delay requirements.

        \paragraph{Service Channel Conditions}
	The channel states are given by a wireless 2.4GHz dataset \cite{channeldata}, which samples the received signal strength indicator (RSSI) in the check-in hall of Shenzhen Baoan International Airport.
	Each channel state is quantized into 4 states, each representing a RSSI level, which is visualized in Figure \ref{subfig:hmchannel}, whose average channel states are given in Table \ref{tab:info}.
        \vspace{-.4cm}
	\begin{table}[H]
		\caption{Summary of arrivals and channel states.}
		\vspace{-.3cm}
		\label{tab:info}
		\begin{tabular}{cllll}
			\toprule
			User & Rate & $\tau_i$ & Character & Avg. Channel\\
			\midrule
			1 & 1.96 & 6 & File Transmission & 1.79 \\
			2 & 0.91 & 6 & Online Forum & 1.83 \\
			3 & 2.46 & 1 & VR Gaming & 1.82 \\
			4 & 0.70 & 1 & Text Communication & 1.77 \\
			\bottomrule
		\end{tabular}
	\end{table}
        \vspace{-.5cm}


	
	\paragraph{Service Outcomes}
	The probability that a transmission for user $i$ is successful under channel state $c$ and allocated resource $e$ is
	\begin{equation}
		P_i(e,c)=\frac{2}{1+\exp[-{2e}/(f_i^{3}c)]}-1
	\end{equation}
	as that in \cite{chen2018timely}, where $f_i$ denotes the distance between user $i$ and the server.
	This models a wireless downlink system.

	\subsection{Ablation Study}
	We first present an ablation study on network architectures and important hyperparameters including policy update frequency and learning rate in this subsection. The results will guide us in our algorithm parameter setting. To keep the underlying MDP well-defined with time-invariant distributions, in the ablation study, we use simulated arrivals and channels in the environment by pre-specifying their distributions. 
	
	\subsubsection{Network Architectures}\label{subsubsec:na}
\begin{figure}[htbp]
	    \vspace{-.5cm}
		\centering
		\subfigure[Single-branch with $a_{t-1}$\label{subfig:sb}]{
			\includegraphics[height=3.8cm]{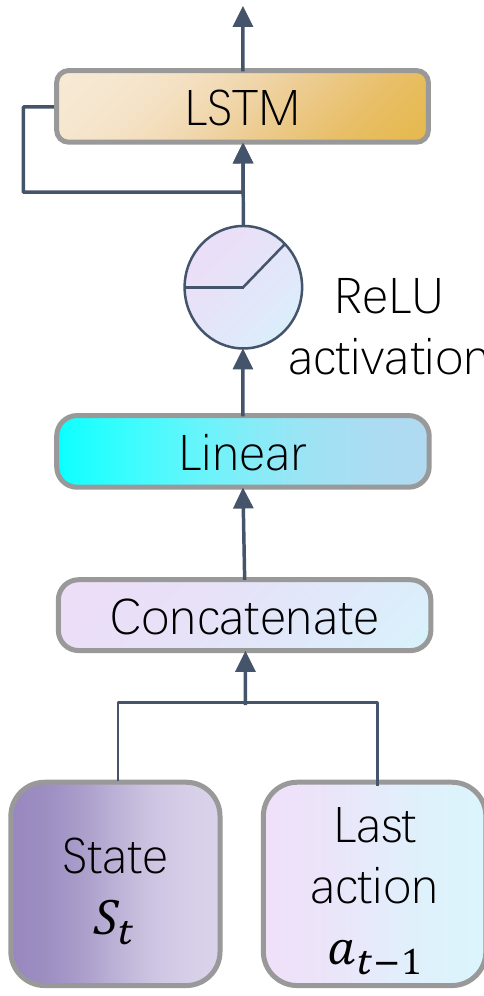}
		}
		\subfigure[Double-branch without $a_{t-1}$\label{subfig:dbnl}]{
			\includegraphics[height=3.5cm]{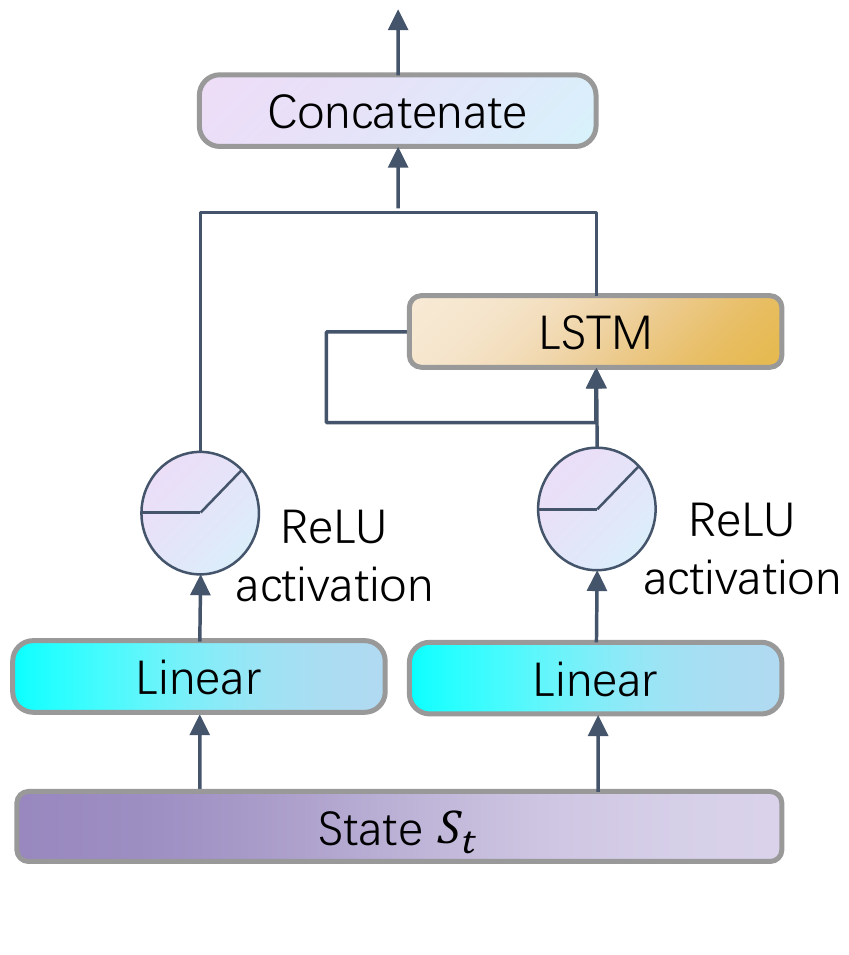}
		}
		\subfigure[Single-branch without $a_{t-1}$\label{subfig:sbnl}]{
			\includegraphics[height=2.9cm]{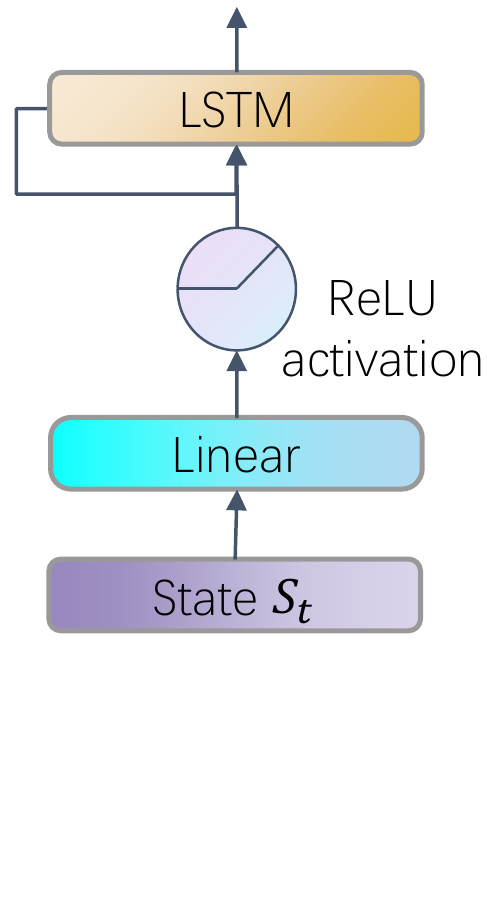}
		}
		\vspace{-.5cm}
		\caption{Different Network Architectures (part).}
		\label{fig:asna}
		\vspace{-.5cm}
	\end{figure}

\begin{figure}[htbp]
		\centering
		\hspace{-1cm}
		\subfigure[Different network architecture\label{subfig:nabar}]{
			\includegraphics[width=.5\columnwidth]{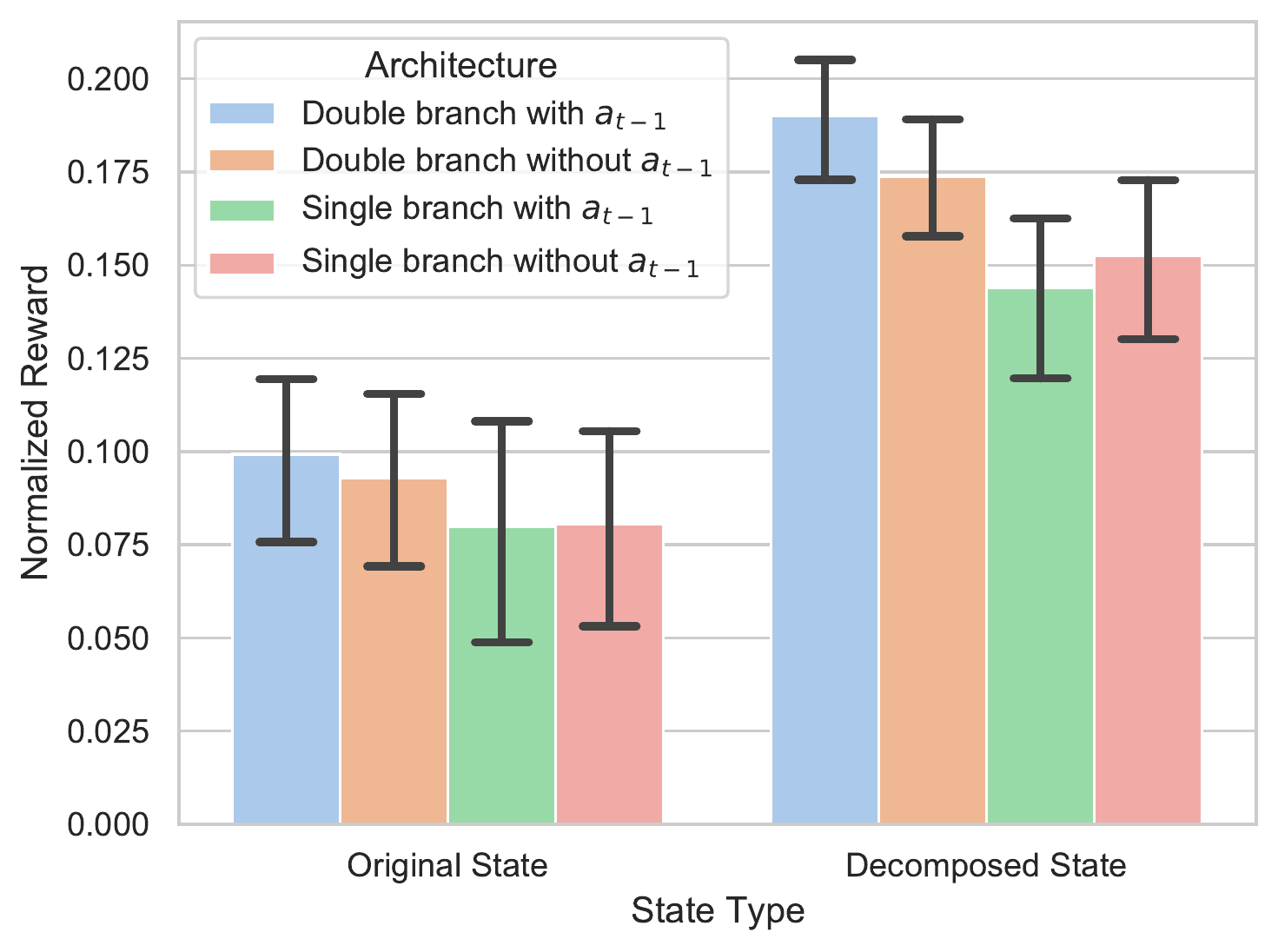}
		}
		\hspace{-.5cm}
		\subfigure[Different policy frequencies\label{subfig:pfbar}]{
			\includegraphics[width=.5\columnwidth]{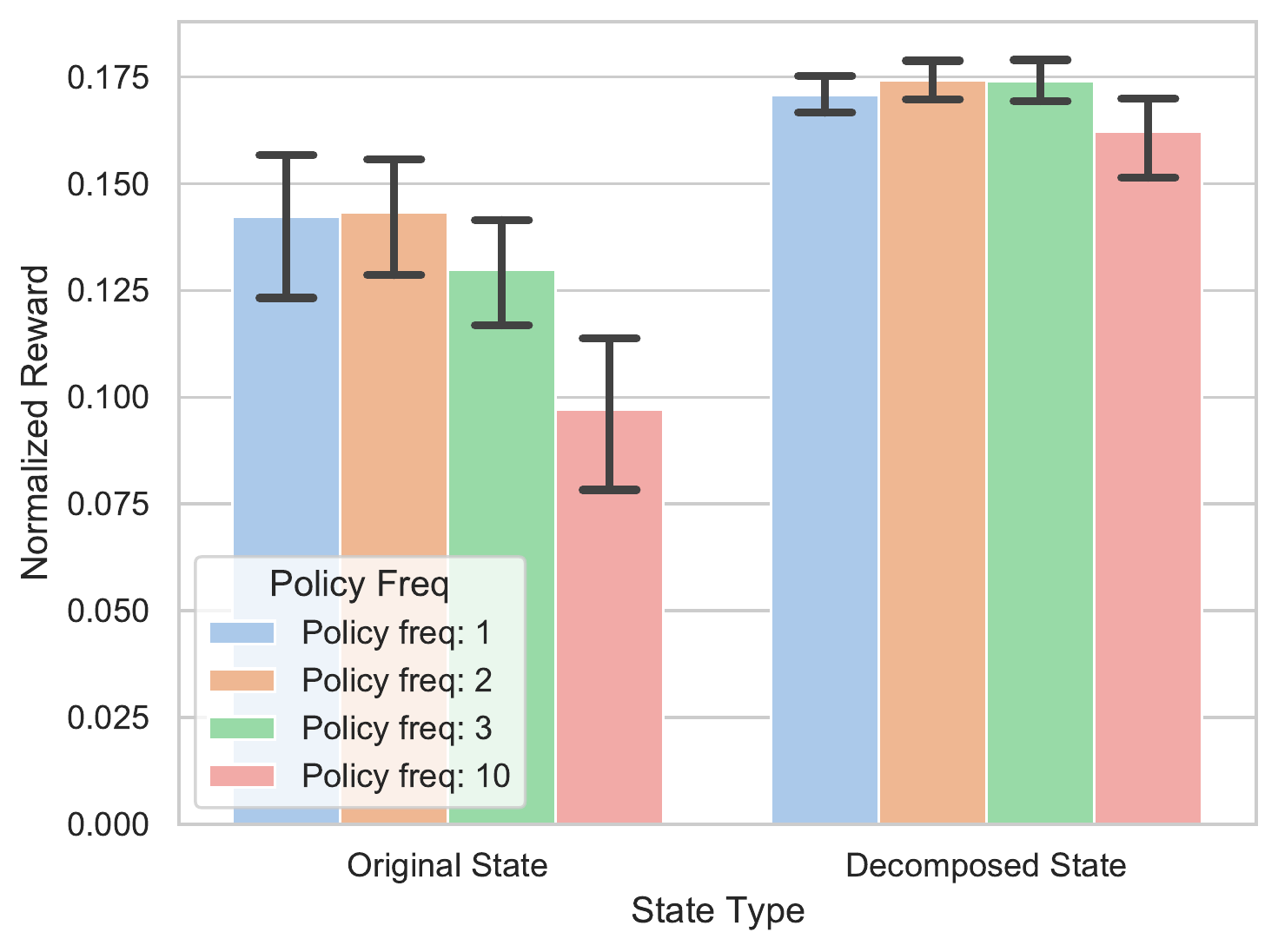}
		}
		\vspace{-.5cm}
		\caption{Experiments of the ablation study.}
		\label{fig:napf}
		\vspace{-.6cm}
	\end{figure}

	We compare different network architectures in Figure \ref{fig:asna} with our architecture in Figure \ref{fig:critic}.\footnote{We only present the first several layers of the critic network, while the last two layers are the same as that in Figure \ref{fig:critic}.} 
	These architectures are tested with $\mathtt{RSD4}$ on the task of maximizing the Lagrangian function in Eq. (\ref{eq:lg}) on $\lambda=\{0,0.1,...,1\}$, since we find this range of $\lambda$ is the region of possible optimal multiplier $\lambda^*$  for our environment setting.
	We normalize the rewards obtained under different $\lambda$ and results are shown in Figure \ref{subfig:nabar}, where rewards obtained with user-level decomposition mentioned in Section \ref{subsec:ds} are presented as well.
	We find that double-branch architecture obtained maximum rewards in both situations, and rewards are further improved by taking the last action $a_{t-1}$ as input to LSTM (see Figure \ref{fig:asna}).
	This validates the advantage of our proposed double-branch architecture and implies that taking action $a_{t-1}$ in the last timestep into account can better capture hidden correlation across time. 

        \vspace{-.2cm}
	
	\subsubsection{Policy Delay}
	
	$\mathtt{RSD4}$ adopts a delayed policy update mechanism (line $25$ in Algorithm \ref{alg:rsd4}), to avoid training divergence due to overestimating a poor policy  \cite{fujimoto2018addressing}.
	The policy network is updated at a lower frequency than the value network to minimize error before introducing a policy update.
	We evaluate different policy frequencies on the same simulated environment as Section \ref{subsubsec:na} in Figure \ref{subfig:pfbar}. The results show that a moderate policy frequency of $2$ improves peak reward in both original and decomposed cases.
	
	\subsubsection{Learning Rate}
	
\begin{figure}[htbp]
		\centering
		\hspace{-.5cm}
		\subfigure[$\lambda_{k}$\label{subfig:20lamb}]{
			\includegraphics[width=.5\columnwidth]{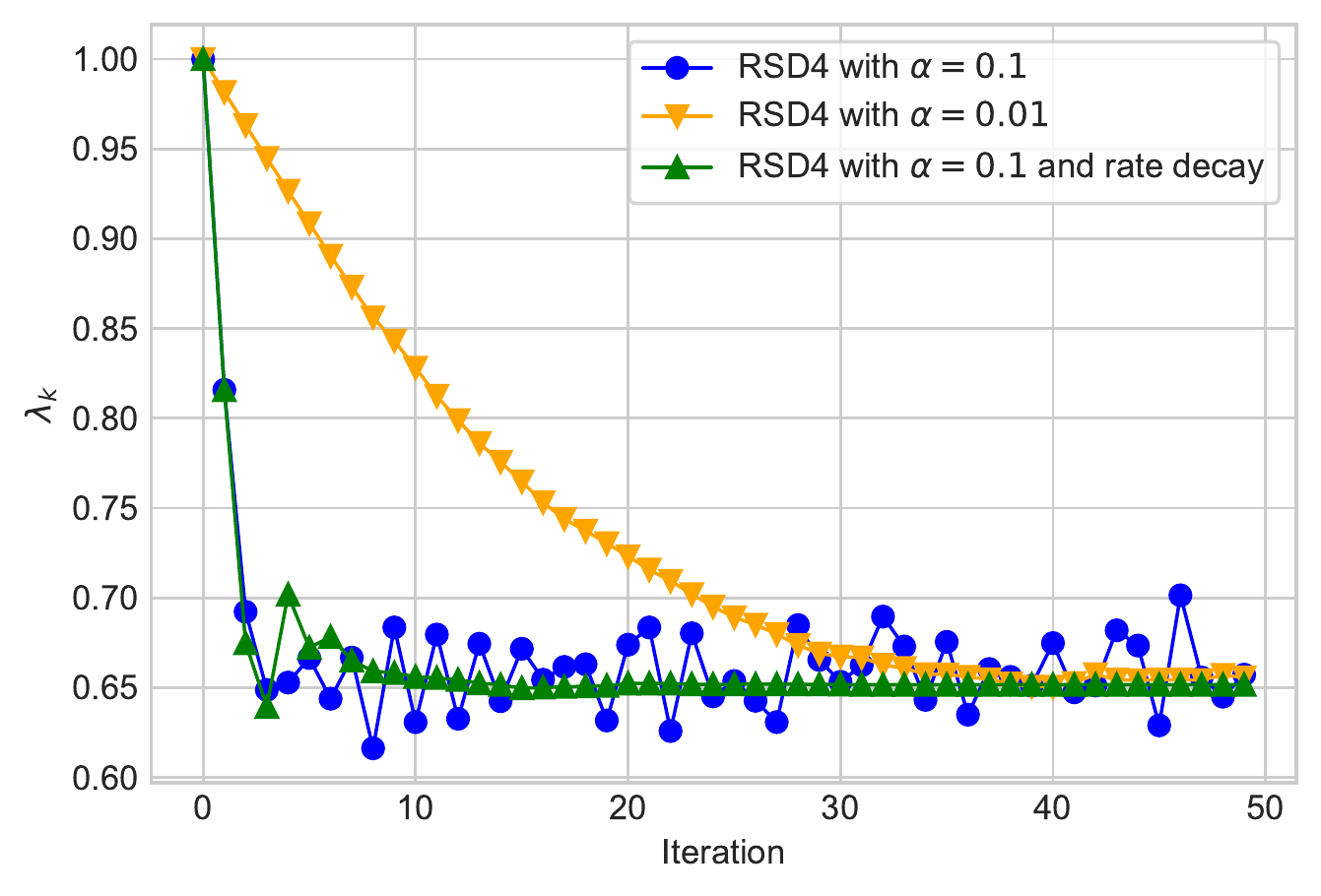}
		}
		\hspace{-.5cm}
		\subfigure[Resouce consumption\label{subfig:20energy}]{
			\includegraphics[width=.5\columnwidth]{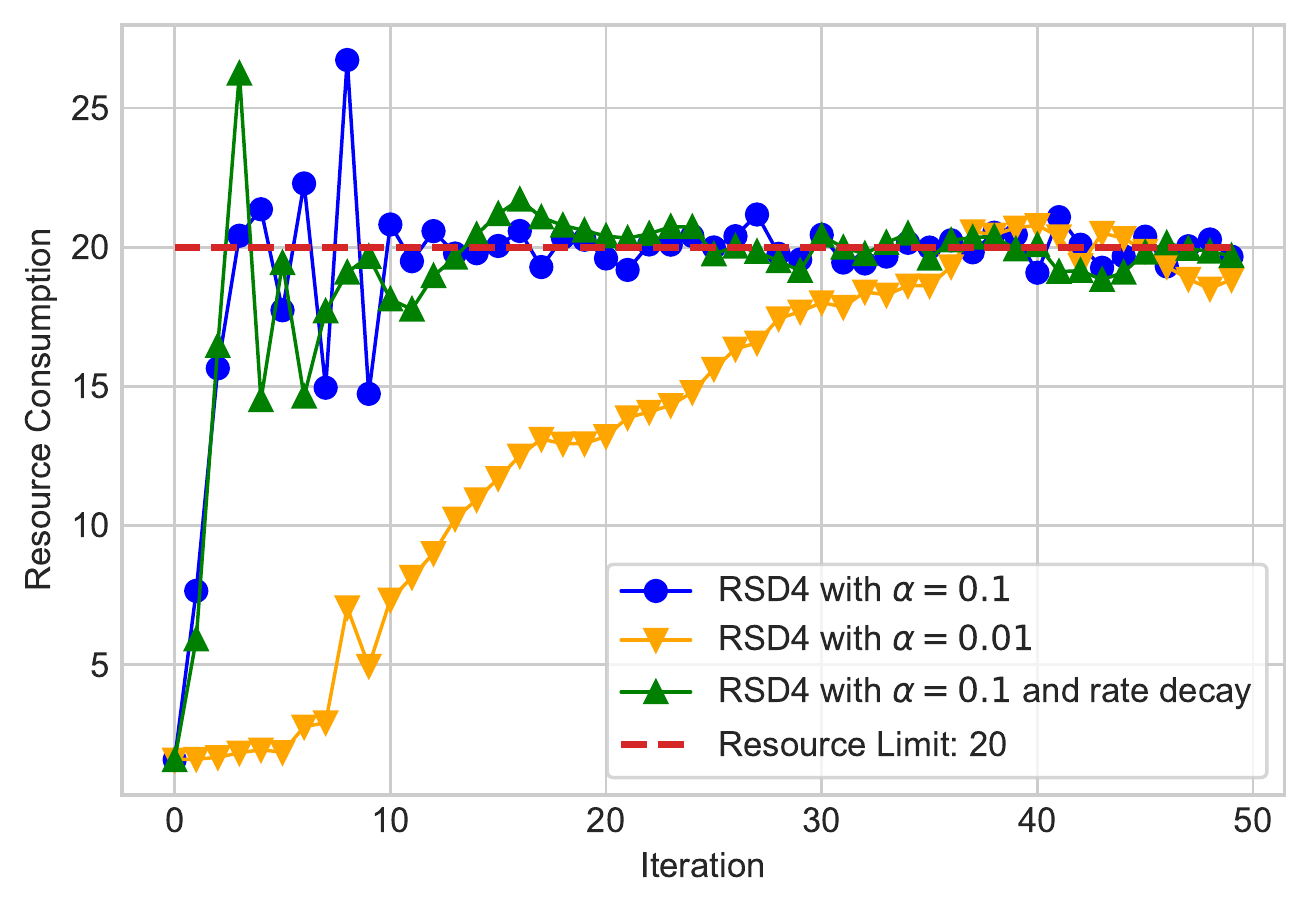}
		}
		\vspace{-.5cm}
		\caption{Iterations under different learning rates.}
		\label{fig:lrtp}
		\vspace{-.6cm}
	\end{figure}

	The learning rate $\alpha$ in Algorithm \ref{alg:rsd4} is critical.
	We execute Algorithm \ref{alg:rsd4} under different $\alpha$ values.  Figure \ref{subfig:20lamb} and \ref{subfig:20lamb} show the iteration of $\lambda_k$ and resource consumptions, respectively.
	A learning rate of $0.1$ is too large such that $\{\lambda_{k}\}$ fluctuates largely, while a small learning rate e.g., $0.01$ does eliminate the large fluctuation of $\{\lambda_{k}\}$ but convergences too slow.
	Thus, we use a decaying learning rate, which halves the learning rate when $\lambda$ flips in three consecutive time slots, i.e., $\alpha_{k+1}=\alpha_k$ if $\lambda_{k+1}\ge\lambda_{k}\ge\lambda_{k-1}$ or $\lambda_{k+1}\le\lambda_{k}\le\lambda_{k-1}$, otherwise $\alpha_{k+1}=\alpha_k/2$. 
	
	\subsection{Performance Comparison}\label{subsec:pc}
	We compare performances of  $\mathtt{RSD4}$ with existing DRL algorithms and classical non-DRL methods. 
	The benchmark DRL algorithms include Twin Delayed DDPG (TD3)  \cite{fujimoto2018addressing} and Softmax DDPG (SD3)  \cite{pan2020softmax}, both are state-of-the-art algorithms. We apply them with Lagrangian dual to ensure average resource constraints. 
	The non-DRL algorithms include Programming,  Uniform, and Earliest Deadline First (EDF) \cite{elsayed2006channel}:
	
	(\romannumeral1) Programming: Solve the following static constrained programming problem $\mathcal{P}_s$ for each time slot $t$ with resource constraint using convex programming  \cite{shen2018fractional} (time index $t$ omitted): 
	\begin{eqnarray}
			\mathcal{P}_s:\quad\max_{\boldsymbol{e}}&&\sum_{i=1}^N\beta_i\sum_{\tau=1}^{\tau_i}B_\tau^iP_i(e_\tau^i, c_{i,\tau})\label{eq:op}\\
		\text{s.t.}&&\sum_{i=1}^N\boldsymbol{e}_i^\top\boldsymbol{B}_i\le E_0\nonumber
	\end{eqnarray} 
	The optimal value $\mathcal{T}_{static}^*$ for $\mathcal{P}_s$ serves as the static optimal throughput for the Problem $\mathcal{P}$ in Eq. (\ref{eq:p1}). 
	
	(\romannumeral2) Uniform: Assign resources to different packets uniformly.
	
		(\romannumeral3) Earliest Deadline First (EDF): Assign all resources to packets in each user's queue with the shortest deadline equally. 
	
	To also evaluate the case when the resource expenditure at each time may be strictly bounded, we consider the hard resource constraint $E_{max}$ for each time slot, i.e., scheduling decisions in each time slot cannot exceed $E_{max}$. We introduce two ways to ensure this.
	The first way is to scale each component equally of $\mathtt{RSD4}$'s decision to ensure the total expenditure is bounded by $E_{max}$ (denoted as $\mathtt{RSD4}$-$E_{max}$-1). The second way is to  allocate resources to packets with smaller remaining time until the total expenditure is $E_{max}$ and set all other allocations to zero (denoted as $\mathtt{RSD4}$-$E_{max}$-2).

	\begin{figure}[htbp]
		\centering
		\hspace{-1cm}
		\subfigure[Throughput\label{subfig:comp_po_throughput}]{
			\includegraphics[width=.6\columnwidth]{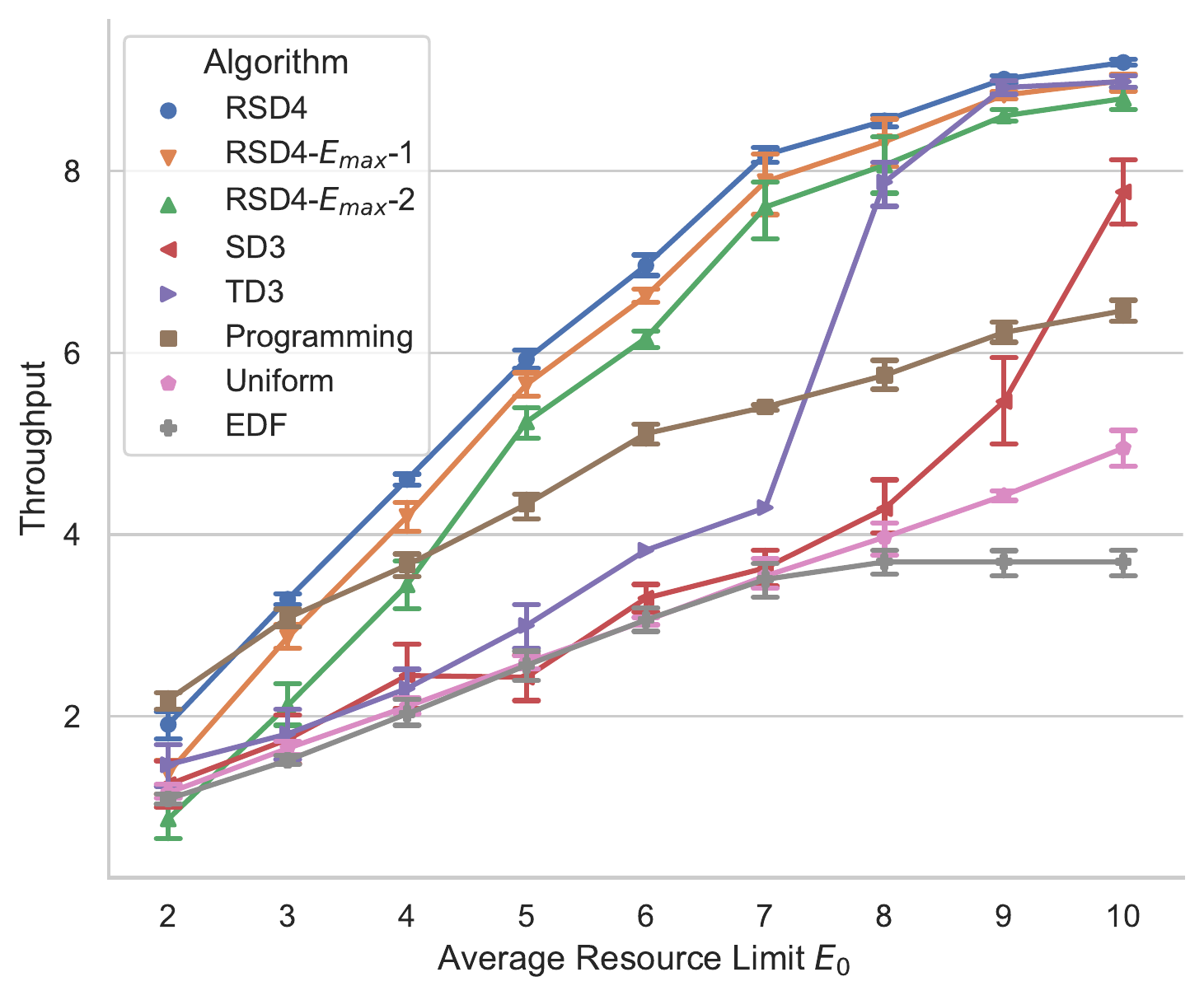}
		}
		\hspace{-.5cm}
		\subfigure[Resource Consumption\label{subfig:comp_po_energy}]{
			\includegraphics[width=.4\columnwidth]{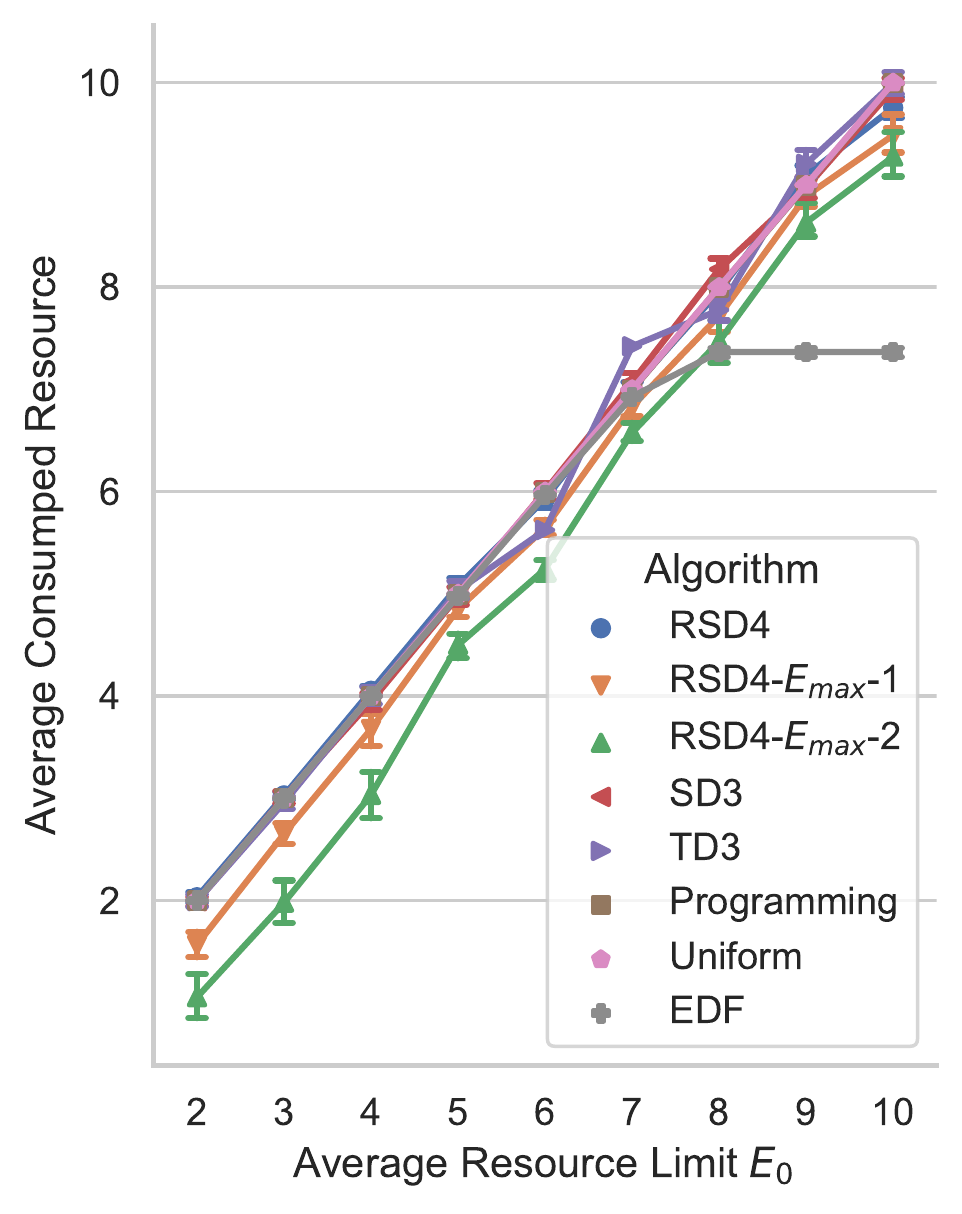}
		}
		\vspace{-.5cm}
		\caption{Comparison of different algorithms.}
		\label{fig:edca}
		\vspace{-.4cm}
	\end{figure}
	
	The comparison of $\mathtt{RSD4}$ with other algorithms is shown in Figure \ref{fig:edca} with $0\le E_0\le 10$, where arrivals and channel conditions are set according to real datasets summarized in Table \ref{tab:info}. Here the channels are assumed to be unobservable to the scheduler. 
	This scenario simulates a multi-user resource-constrained wireless base-station providing delay-constrained scheduling services.
	We find that performances of $\mathtt{RSD4}$-$E_{max}$-1 and $\mathtt{RSD4}$-$E_{max}$-2 are quite similar to $\mathtt{RSD4}$ when setting $E_{max}=2E_0$ in each group of experiments, which means that hard resource constraint can be achieved with our methods without much performance degradation.
	Besides, from Figure \ref{fig:edca}(a),  we see that $\mathtt{RSD4}$ outperforms all benchmarks in each resource limit. 
	In the small-medium  resource regime with $0\le E_0\le7$, the static optimality obtained by Programming ranks only lower than  $\mathtt{RSD4}$. 
	However, in the large resource regime with $8\le E_0\le10$, $\mathtt{RSD4}$ and TD3 outperform classical methods significantly. 
	From Figure \ref{fig:edca}(b), we see that all DRL algorithms satisfy the average resource limit, while  EDF fails  to fully utilize available resources since packets with the shortest deadline are too few to consume all available resources. 
	Results in Figure \ref{fig:edca} illustrate the superiority of $\mathtt{RSD4}$ over the benchmarks. 
	
    
	\subsection{Partially Observable Systems}\label{subsec:pos}
	

    We compare the performance of $\mathtt{RSD4}$ with other DRL algorithms to further validate the effectiveness of its recurrent module.
    Our comparison focuses on maximizing the Lagrangian function in Eq. (\ref{eq:lg}) with the same environment of Section \ref{subsec:pc}, which is essentially equivalent to the task of throughput maximization. 
    In fact,  each average resource limit $E_0$ corresponds to an optimal $\lambda^*$ according to Section \ref{subsec:ld}. Thus, maxmizing the Lagrangian function on some $\lambda$ is equivalent to throughput maximization under some average resource limit of $E_0(\lambda)$ determined by $\lambda$. 
    Thus, focusing on maximizing the Lagrangian function in Eq. (\ref{eq:lg}) allows us to examine a wide range of problems parameterized by $\lambda$. 
	
	
	

	\subsubsection{Missing Buffer State}
	We first consider an extreme environment where the buffer state is unobservable by the agent, and only arrivals and channels are given.
	Specifically, observation $o_t=[\boldsymbol{A}(t),\boldsymbol{c}(t)]$, which implies the agent needs to memorize arrivals and service outcomes across multiple time slots to have accurate estimations of current system states.
	From Figure \ref{subfig:pobuffer}, we find that with buffer state, i.e., the underlying MDP is fully observable, $\mathtt{RSD4}$ and SD3 obtain similar maximum rewards under different $\lambda$. 
	When the buffer state is missing,  $\mathtt{RSD4}$ 
	 still achieves almost the same maximum rewards, whereas  non-recurrent DRL algorithms SD3 and TD3 suffer from significant performance loss. 
	

	\subsubsection{Unobservable Hidden Factors}
\begin{figure}[htbp]
		\centering
		\vspace{-.3cm}
		\hspace{-.5cm}
		\subfigure[Environment without buffer state.\label{subfig:pobuffer}]{
			\includegraphics[width=.5\columnwidth]{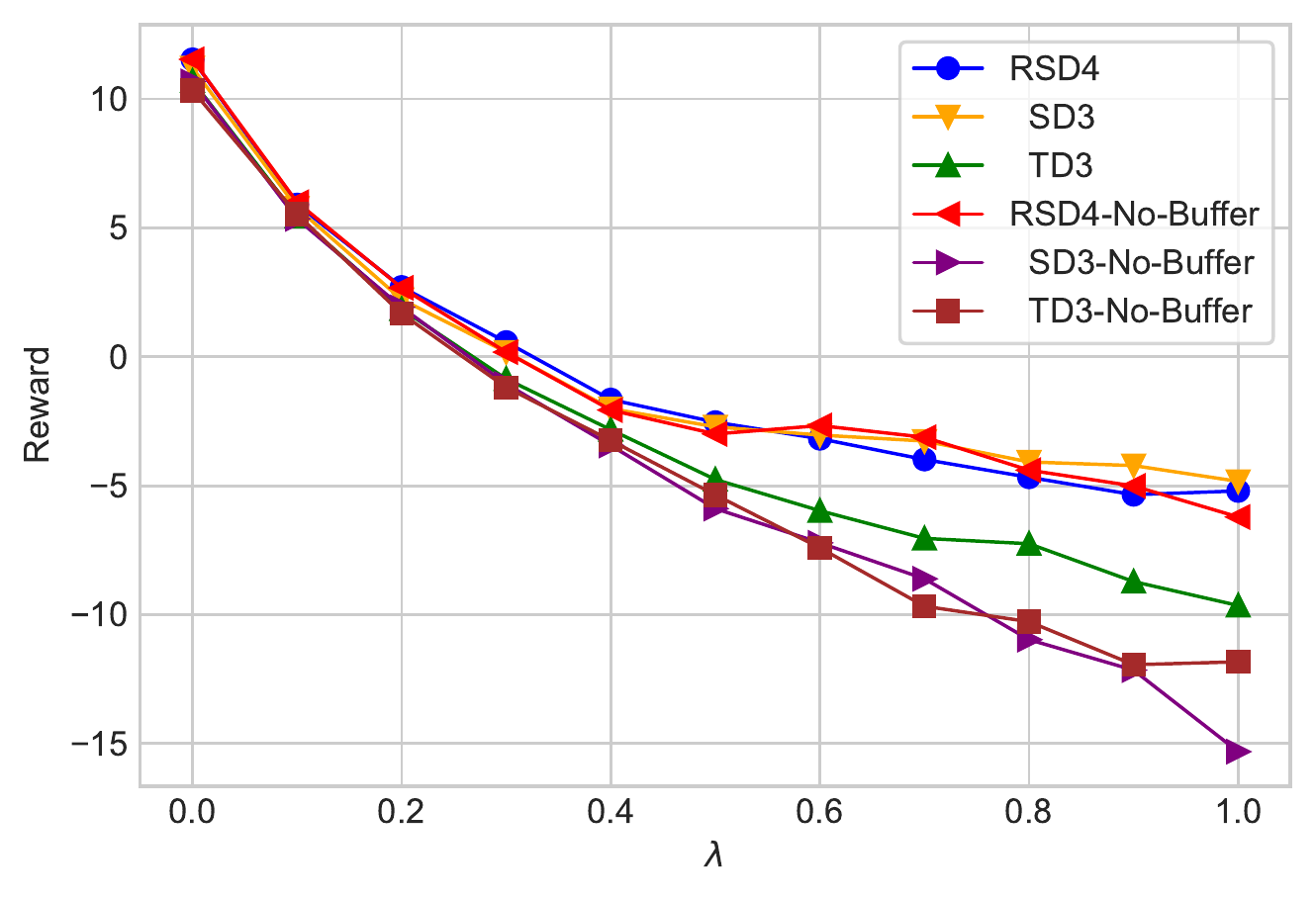}
		}
		\hspace{-.5cm}
		\subfigure[Environment with hidden factors.\label{subfig:pohf}]{
			\includegraphics[width=.5\columnwidth]{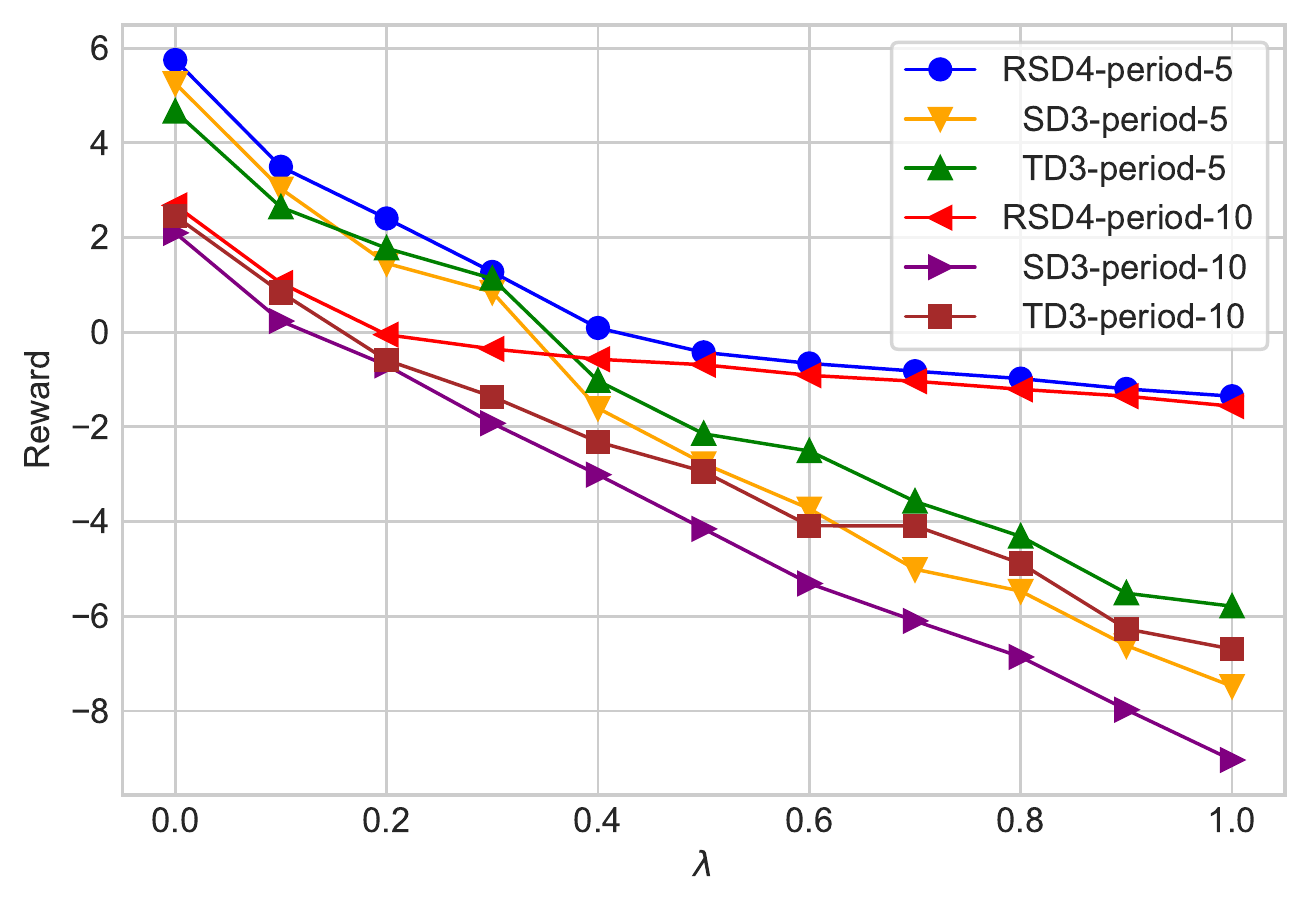}
		}
		\vspace{-.5cm}
		\caption{Various partially observable settings.}
		\vspace{-.5cm}
		\label{fig:vpos}
	\end{figure}

	Another common type of partial observability comes from unobserved hidden factors in the underlying MDP, e.g.,
	vehicle obstruction will influence in-tunnel wireless propagation channel which is hard to trace \cite{song2021impact}.
	We design an environment where service outcomes are related to the time index, i.e., services are available only when the current time slot $t$ is a multiple of some period, e.g., wireless communication interfered by periodic jamming signals. 
	In this case, the underlying MDP is partially observable since the hidden factor, i.e., the period, is unknown to the agent and the observation is $o_t=[\boldsymbol{B}^1(t),...,\boldsymbol{B}^N(t),\boldsymbol{c}(t)]$.
	From Figure \ref{subfig:pohf}, we find that $\mathtt{RSD4}$ outperforms TD3 and SD3 in both tested cases with periods $5$ and $10$, and the performance gain of $\mathtt{RSD4}$ is more significant under large $\lambda$ case, which corresponds to the small resource regime.
	
	\subsubsection{Time-varying Environments}\label{subsubsec:sw}
\begin{figure}[htbp]
    	\centering
    	\vspace{-.3cm}
    	\hspace{-.5cm}
    	\subfigure[Switch on arrival strength\label{subfig:transan}]{
    		\includegraphics[width=.5\columnwidth]{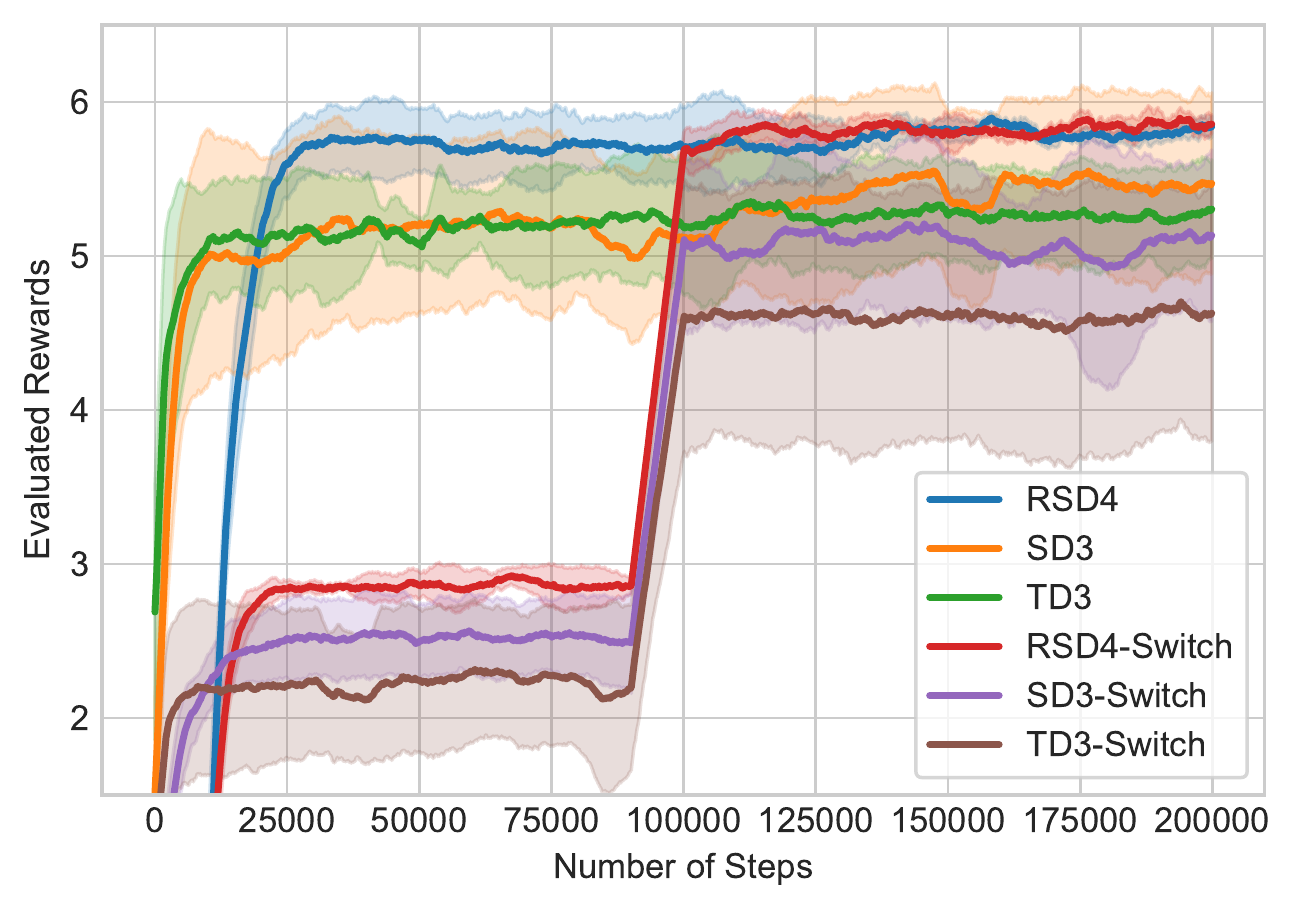}
    	}
    	\hspace{-.5cm}
    	\subfigure[Switch on channel availability\label{subfig:transcs}]{
    		\includegraphics[width=.5\columnwidth]{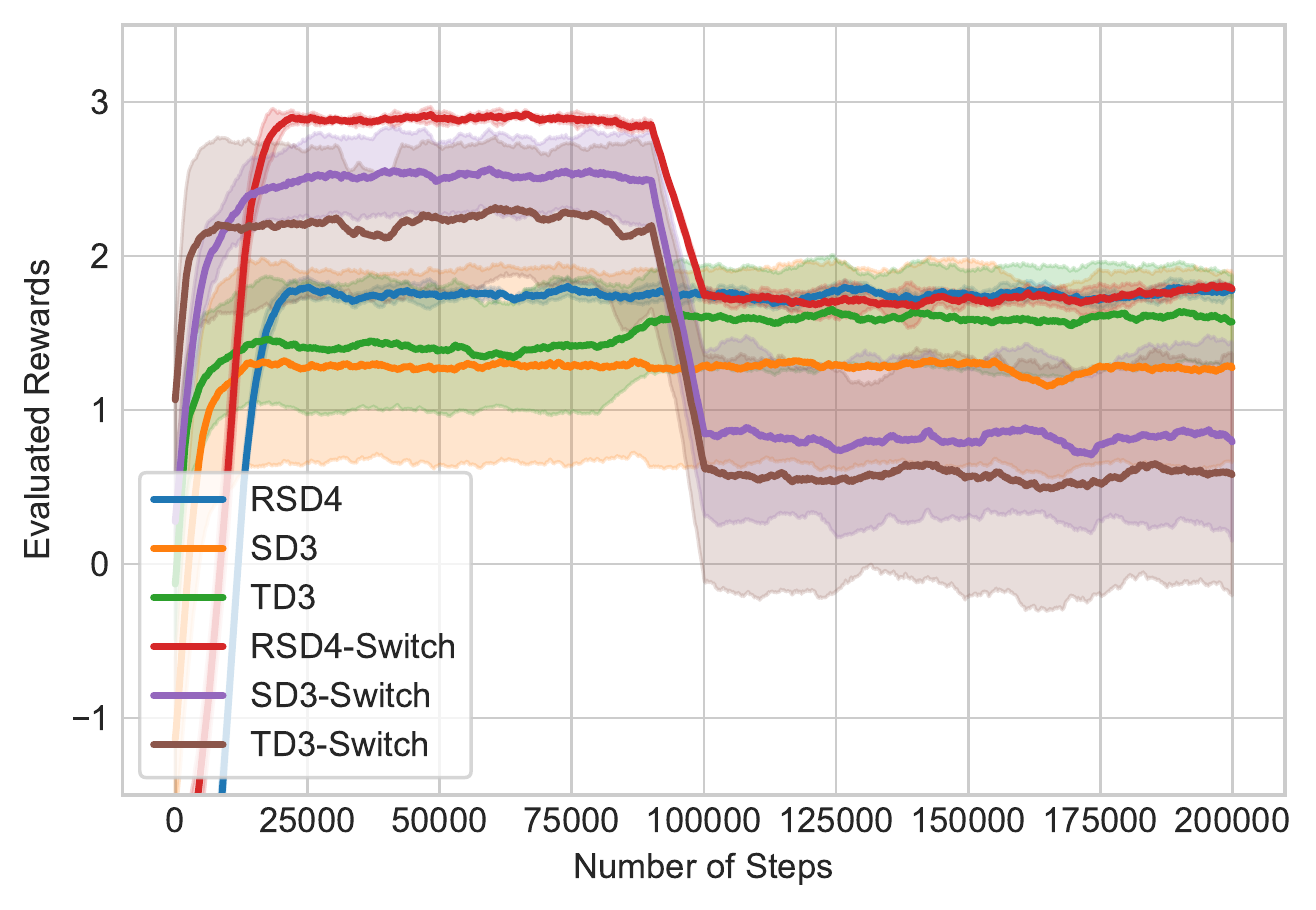}
    	}
    	\vspace{-.5cm}
    	\caption{Experiments on switching environments.}
    	\vspace{-.4cm}
    	\label{fig:switch}
    \end{figure}

	We next investigate 
   partial observability coming from the time variability of underlying system dynamics.
    We design two types of time variability by switching environment dynamics during the experiments, i.e., in one setting we double the rate of arrivals and in the other we change the channel statistics. In both cases the channel states are unobservable.  
    The switch happens at slot $100,000$ and results are shown in Figure \ref{fig:switch} for the task of maximizing Lagrangian function under $\lambda=0.5$, where the evaluated rewards of training processes under  the environments with switching in the middle and switching at the very beginning.
    In both cases, we find that, soon after switching,  $\mathtt{RSD4}$ reaches the same rewards one can  obtain by training in these  environments from the very beginning, while TD3 and SD3 both suffer from  reward loss after switching, even if they are trained from the beginning for this new environment. This validates the robustness and performance of $\mathtt{RSD4}$ in time-varying environments.
    %

    \begin{remark}
	    Experimental results on these three partially observable environments show the superiority of $\mathtt{RSD4}$ at addressing partial observability issues. They also show the benefits of adopting a POMDP formulation in practical scheduling problems, since partial observability can arise from many sources. This comparison also sheds light on understanding the sub-optimality of non-recurrent DRL algorithms or classical non-DRL methods in POMDP. 
	\end{remark}
	
	\subsection{Scalability by Decomposition and Merging}\label{subsec:scalability}
	We now turn to investigate the scalability of the algorithm by conducting experiments in systems with a large number of users and with multi-hop structures. We focus on the setting where the channel states are assumed to be unobservable.
	We will show that the user-level decomposition and node-level techniques in $\mathtt{RSD4}$ make it a highly scalable solution. 
	
	
	\subsubsection{Large-Scale System}
\begin{figure}[htbp]
        \vspace{-.4cm}
		\centering
		\hspace{-.5cm}
		\subfigure[Rewards on different user scales.\label{subfig:sa_reward}]{
			\includegraphics[width=.4\linewidth]{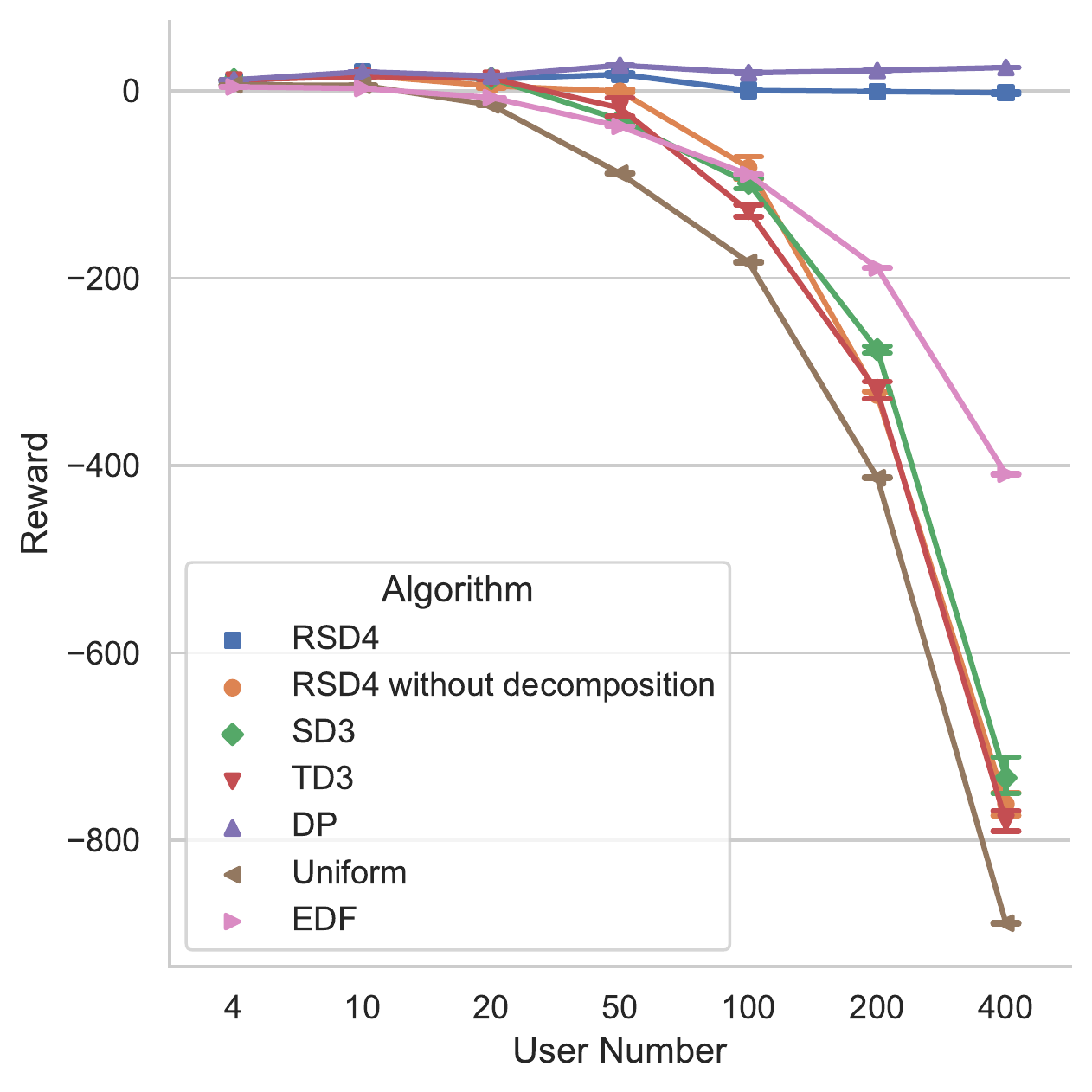}
		}
		\hspace{-.5cm}
		\subfigure[Training process with 50 users.\label{subfig:train50}]{
			\includegraphics[width=.6\linewidth]{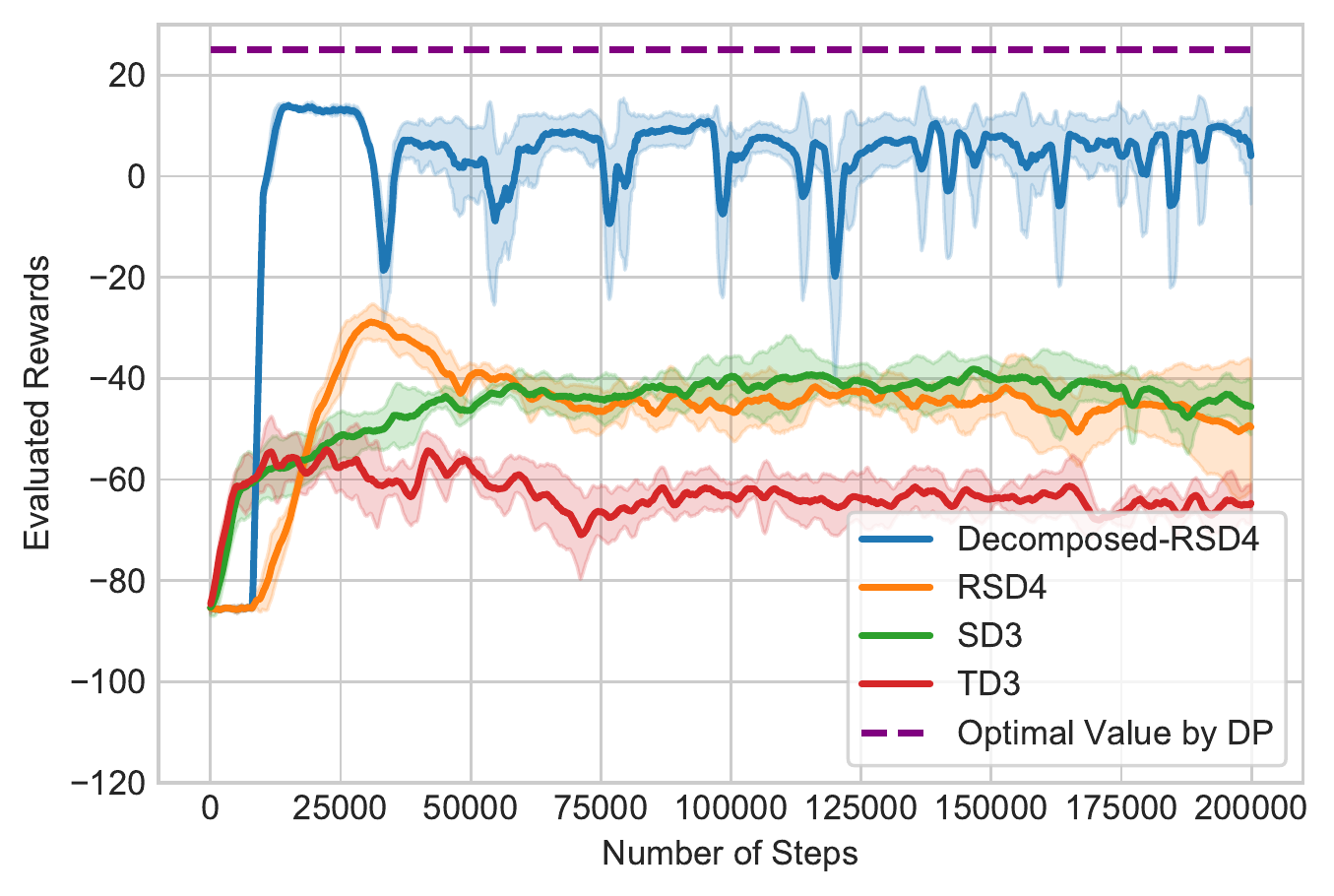}
		}
		\vspace{-.5cm}
		\caption{Experiments on scalability.}
		\label{fig:sa}
		\vspace{-.3cm}
	\end{figure}

	We present the evaluated rewards of $\mathtt{RSD4}$ with other algorithms on the task of maximizing the Lagrangian function with $\lambda=0.3$ on different numbers of users ranging from $4$ users to $400$ users in Figure \ref{subfig:sa_reward}.
	Here we generate arrivals and channels according to predefined random processes, so that system dynamics are known and we can get the optimal reward by DP. 
    It is worth noting that computing the optimal induces extremely high computation overhead in large-scale tasks and requires accessing full system dynamics, and is not practical in real-world applications. 
    We examine this case here emphasize the near-optimality and effectiveness of $\mathtt{RSD4}$. 
    All baselines are implemented with the same set of hyperparameters to ensure fairness.
    
	From Figure \ref{subfig:sa_reward}, we find that when user number is less than or equals to $20$, the performance of different baselines do not differ much, and all of them achieve near-optimal rewards. 
	However, with an increasing number of users, all DRL algorithms, including $\mathtt{RSD4}$ without decomposition fail, as well as  Uniform and EDF. 
	Besides, $\mathtt{RSD4}$ always achieves the best and near-optimal performance regardless of the number of users. 
	Figure \ref{subfig:train50} also shows the learning curve under $50$ users (from which the performance of different baselines start to diverge significantly), where only  $\mathtt{RSD4}$ achieves the nearly optimal reward. We also see that $\mathtt{RSD4}$ without state-decomposition fails in this case. 
    This demonstrates the necessity of state-decomposition, without which the algorithm parameter amount will become too large and prohibits efficient training.
	
	
	\begin{remark}
	    When the system scale is beyond the hypothesis dimension of the underlying neural network, e.g., when the user number is larger than $20$ in Figure \ref{subfig:sa_reward}, the performance of DRL algorithms will degrade rapidly, which means neural networks with more hyperparameters will be required.
	    However, with more hyperparameters, neural networks are much harder to train and require significantly more computational power. 
	    With state decomposition, in contrast, $\mathtt{RSD4}$ efficiently controls the state dimension and still retains the near-optimal performance. 
	    This validates the effectiveness of our proposed user-level decomposition. 
	\end{remark}
    
	\subsubsection{Multi-hop Network}
\begin{figure}[htbp]
		\centering
		\vspace{-.5cm}
		\hspace{-.5cm}
		\subfigure[Throughput of different algorithms $\quad\quad$ on a multi-hop network with 12 flows.\label{subfig:networkbar}]{
			\includegraphics[width=.5\linewidth]{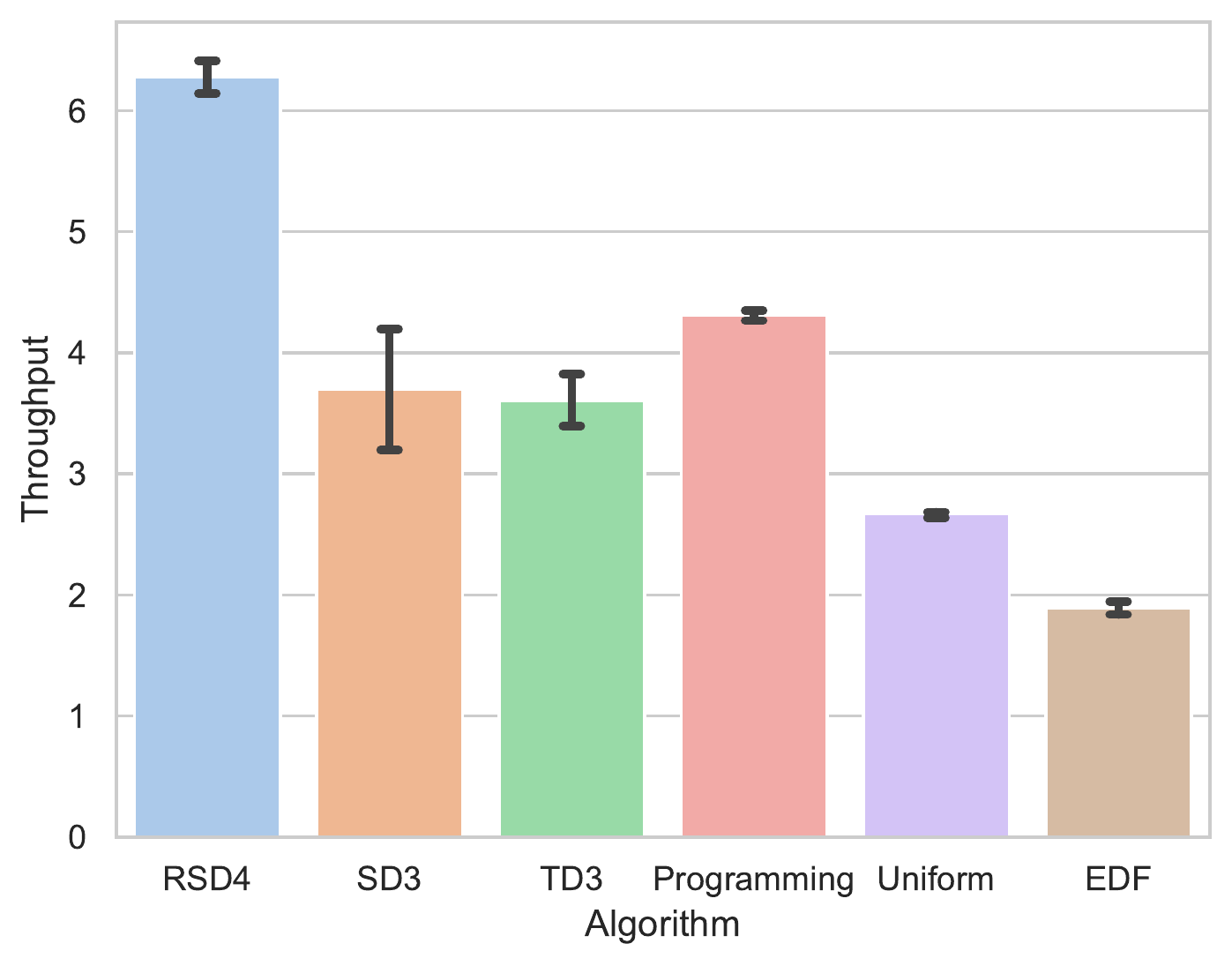}
		}
		\hspace{-.5cm}
		\subfigure[Rewards on different user scales of multi-hop networks.\label{subfig:expmultihop}]{
			\includegraphics[width=.5\linewidth]{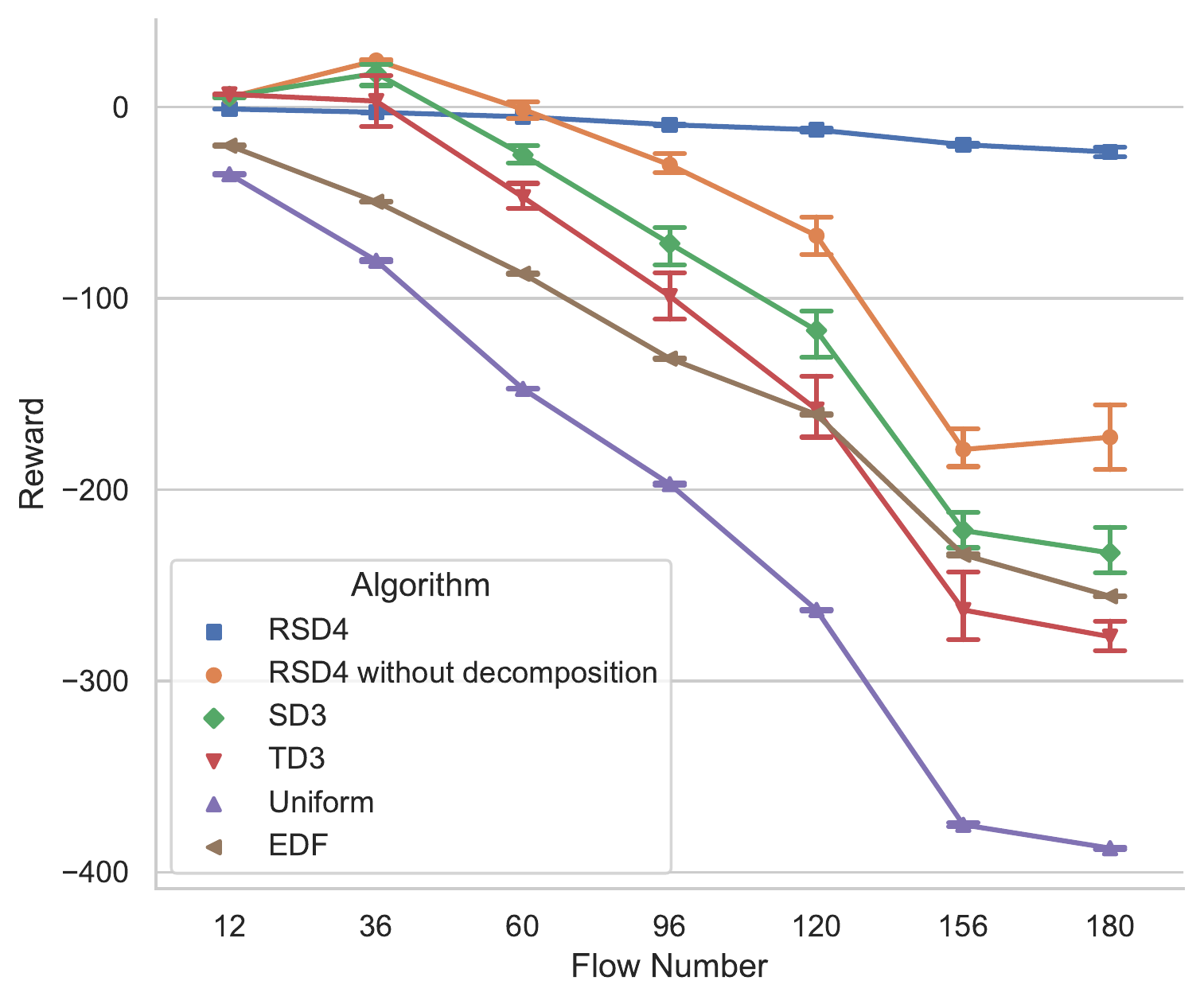}
		}
		\vspace{-.5cm}
		\caption{Experiments on multihop networks.}
		\label{fig:nm}
		\vspace{-.5cm}
	\end{figure}

	We now turn to the multihop network setting.
    Recall that we propose node-level merging in Section \ref{subsec:multihop} to schedule under multihop networks.
	We conduct experiments on a multi-hop network depicted in Figure  \ref{subfig:network}, where there are three paths, $Path_1=1\rightarrow2\rightarrow3\rightarrow5$, $Path_2=2\rightarrow4\rightarrow6$,  and $Path_3=2\rightarrow3$. $Path_1$, $Path_2$, and $Path_3$ has $3$, $4$, and $5$ flows passing through respectively. The $12$ flows have different deadlines ranging from $3$ to $5$ timeslots and the scheduler needs to make decisions at nodes $1$, $2$, $3$ and $4$ simultaneously, with average resource limits  $10$, $30$, $0.3$, $3$, respectively.
	The arrival and channel states are drawn from real-world datasets specified above, and the throughput obtained by different algorithms are shown in Figure \ref{subfig:networkbar}. 
	
	We see that $\mathtt{RSD4}$ achieves the maximum throughput and significantly outperforms other benchmarks, including TD3 and SD3 and non-DRL-based methods Programming, Uniform, and EDF.
	To further validate the scalability of our $\mathtt{RSD4}$ algorithm, we fix the four multiplier values to $0.5$ and 
	compare it with other benchmarks on the task of maximizing the Lagrangian function for this multi-hop network. 
	We  test the algorithm with an increasing number of flows from $12$ flows to  $180$ flows. 
	$\mathtt{RSD4}$  shows a larger performance gain over other algorithms as the number of flows increase and saves computation resources by unified training.
	
	\section{Conclusion}

	This paper studies the problem of multi-user  latency-constrained scheduling with average resource constraints. 
	To tackle partial observability and scalability issues, we propose a novel DRL algorithm, $\mathtt{RSD4}$, which is a data-driven method based on a POMDP formulation. $\mathtt{RSD4}$ guarantees resource and delay constraints by Lagrangian dual and delay-sensitive queues, respectively. 
	$\mathtt{RSD4}$ successfully tackles partially observable issues with a recurrent network module. 
	It also enables robust value estimation with the softmax operator, and introduces the user-level decomposition and node-level merging techniques to ensure scalability. 
	We conduct extensive experiments on both simulated environments and real-world datasets. Our results show that $\mathtt{RSD4}$ is robust to various system dynamics and partially observable settings, and significantly outperforms existing DRL/non-DRL-based benchmarks.
\section*{Acknowledgements}

This work is supported in part by the Technology and Innovation Major Project of the Ministry of Science and Technology of China under Grant 2020AAA0108400 and 2020AAA0108403, the Tsinghua University Initiative Scientific Research Program, and Tsinghua Precision Medicine Foundation 10001020109.
\bibliographystyle{ACM-Reference-Format}
\bibliography{sample-base}

\end{document}